\documentclass[journal]{IEEEtran}
\usepackage{xurl} 
\usepackage{hyperref}
\usepackage{fix-cm}
\usepackage{amsmath,amsfonts}
\usepackage{array}
\usepackage[noadjust]{cite}
\usepackage{textcomp}
\usepackage{stfloats}
\usepackage{url}
\usepackage{verbatim}
\usepackage{graphicx}
\def\BibTeX{{\rm B\kern-.05em{\sc i\kern-.025em b}\kern-.08em
    T\kern-.1667em\lower.7ex\hbox{E}\kern-.125emX}}
\usepackage{balance}
\usepackage{tikz}
\usetikzlibrary{fadings}
\usetikzlibrary{patterns}
\usetikzlibrary{shadows.blur}
\usetikzlibrary{shapes}   
\usepackage{siunitx}
\DeclareSIUnit\mt{\milli\tesla} 
\usepackage[utf8]{inputenc}
\usepackage[english]{babel}
\usepackage{xurl} 
\usepackage[linesnumbered,ruled,lined]{algorithm2e}
\usepackage{balance}
\usepackage{multirow} 
\usepackage{subcaption}
\usepackage{tikz}
\usetikzlibrary{arrows.meta, positioning, fit}
\usepackage{algcompatible}
\usepackage{adjustbox}
\usepackage{color}

\usepackage{pgfplots}
\usepackage{tikz}
\usepackage{bbm}
\let\oldnl\nl
\newcommand{\nonl}{\renewcommand{\nl}{\let\nl\oldnl}}
\usepackage{mathrsfs}

\usepackage{graphicx}  
\usetikzlibrary{positioning, fit, arrows.meta, shapes}

\newcolumntype{L}[1]{>{\raggedright\let\newline\\\arraybackslash\hspace{0pt}}m{#1}}
\newcolumntype{C}[1]{>{\centering\let\newline\\\arraybackslash\hspace{0pt}}m{#1}}
\newcolumntype{R}[1]{>{\raggedleft\let\newline\\\arraybackslash\hspace{0pt}}m{#1}}

\algnewcommand\KwEvl{\textbf{Evaluation:}}
\usepackage{booktabs}

\usepackage{makecell}
\usepackage{amsmath}
\usepackage{siunitx}
\usepackage{caption} 
\usepackage{etoolbox}
\usepackage{tikz}

\newrobustcmd*{\mysquare}[1]{\tikz{\filldraw[draw=#1,fill=#1] (0,0)
rectangle (0.2cm,0.2cm);}}

\newrobustcmd*{\mycircle}[1]{\tikz{\filldraw[draw=#1,fill=#1] (0,0) circle [radius=0.1cm];}}

\newrobustcmd*{\mytriangle}[1]{\tikz{\filldraw[draw=#1,fill=#1] (0,0) --
(0.2cm,0) -- (0.1cm,0.2cm);}}

\usepackage{caption}

\usepackage{titlesec}
\titlespacing{\section}{0pt}{*.65}{*.65}
\titlespacing{\subsection}{0pt}{*.65}{*.65}
\titlespacing{\subsubsection}{0pt}{*0}{*0}

\usepackage{pgfplots}
\pgfplotsset{compat=1.18}

\begin{document}
\title{Causal Discovery-Driven Secure mmWave Beam Alignment with Defensive Variational Autoencoders}

\title{Causal AI-Driven Secure mmWave Beam Alignment \\ with  Defensive Variational Autoencoders}
\author{Nasir~Khan,~\IEEEmembership{Graduate Student Member,~IEEE}, Asmaa~Abdallah,~\IEEEmembership{Senior~Member,~IEEE}, \\ Abdulkadir~Celik,~\IEEEmembership{Senior~Member,~IEEE},~Ahmed~M.~Eltawil,~\IEEEmembership{Senior Member,~IEEE},\\ Sinem Coleri,~\IEEEmembership{Fellow,~IEEE}, and~Cicek~Cavdar,~\IEEEmembership{Member,~IEEE}

\thanks{Nasir khan and Cicek Cavdar are with the School of Electrical
Engineering and Computer Science, KTH Royal Institute of Technology,
100 44 Stockholm, Sweden (e-mail: $\lbrace$nkha, cavdar$\rbrace$@kth.se)}
 \thanks{Asmaa Abdallah and Ahmed M. Eltawil are with the Computer, Electrical, and Mathematical Sciences and Engineering Division, King Abdullah University of Science and Technology, Thuwal 23955, Saudi Arabia (e-mail: $\lbrace$asmaa.abdallah, ahmed.eltawil$\rbrace$@kaust.edu.sa)}
 \thanks{Abdulkadir Celik is with the School of Electronics and Computer Science, University of Southampton, Southampton SO17 1BJ, U.K. (e-mail:a.celik@soton.ac.uk)}
\thanks{Sinem Coleri is with the department of Electrical and Electronics Engineering, Koc University, Istanbul, Turkey, email: (scoleri@ku.edu.tr). 

This work is supported in part by the project ``Towards Resilient 6G Networks: A Swedish-Finnish Joint Undertaking'' (FiSRE), funded by Vinnova in Sweden, under Grant No.\ 2026-00530, and in part by the CELTIC-NEXT project ``Robust and AI Native 6G for Green Networks'' (RAI-6GREEN), funded by Vinnova in Sweden, under Project ID C2023/1-9 and Grant No.\ 2023-02534. The work of Sinem Coleri is supported by the Scientific and Technological Research Council of Turkey (TUBITAK) under Grant No.\ 119C058 and by Ford Otosan.} }

\maketitle
\begin{abstract}  
Efficient and secure beam alignment is essential for mmWave communication systems. However, existing deep learning (DL)–based beam alignment methods often neglect causal relationships between inputs and outputs, resulting in limited interpretability, poor generalization, and excessive beam sweeping. Moreover, these DL solutions are vulnerable to adversarial attacks, where small perturbations in received power measurements can mislead beam classifiers.
To overcome these issues, we propose an innovative causally-aware DL framework integrating causal discovery into the beam alignment process. A two-stage causal beam selection algorithm first learns a Bayesian graph capturing dependencies between received signal strength indicators (RSSI) as inputs and the optimal beam index as the target variable, then uses it to select a minimal set of relevant  input features for the DL classifier. To mitigate adversarial attacks, we introduce a defensive variational autoencoder (DVAE) that maps potentially perturbed inputs back to their clean versions. We show that the end-to-end fine-tuning of the downstream beam classifier on the DVAE-reconstructed RSSI representations further boosts prediction accuracy. 
Simulation results show that causal beam selection achieves accuracy comparable to conventional methods while providing an \(\mathbf{829\times}\) lower per-sample computational cost than SHAP (Shapley Additive Explanations), a widely used explainable AI(XAI)-based feature selection method. It further reduces beam-sweeping overhead by \(\mathbf{59.4\%}\) and \(\mathbf{89.8\%}\), compared to conventional DL-based beam selection and exhaustive search, respectively. The DVAE-based defense effectively denoises adversarial inputs, sustaining 85\% accuracy across different perturbation bounds,  retains baseline performance on unperturbed RSSI inputs and surpasses defensive distillation defense scheme by 15--20\% on average under different attack scenarios.

\end{abstract}


\IEEEpeerreviewmaketitle

\section{Introduction}
\IEEEPARstart{T}{he} abundant spectrum available at millimeter-wave (mmWave) frequencies offers a promising means to support extremely high data rates for the existing  $5^{\text{th}}$ Generation (5G)
wireless networks and  is expected to play a central role in future $6^{\text{th}}$ Generation (6G) technologies. However, the severe path and penetration losses inherent at these frequencies necessitate the use of highly directional narrow beamforming to effectively focus the transmitted energy and maintain reliable communication links. While narrow beamforming enhances directionality and gain, it also increases spatial selectivity, making the system more susceptible to beam misalignment. This challenge becomes even more pronounced in future systems that are expected to employ larger antenna arrays and narrower beams to achieve higher beamforming performance.  Therefore, efficient beam management (BM) encompassing procedures such as initial beam alignment, beam tracking, and recovery from beam failures is vital for ensuring reliable operation in future wireless communication systems \cite{IA}. Beam alignment is carried out during initial access (IA) and it
becomes increasingly difficult when narrower beams are used.

Conventional beam alignment schemes select the optimal beam from a predefined set of candidate narrow beams within a codebook via over-the-air measurements between the base station (BS) and the user equipment (UE) feedback \cite{tutorialBA}.  For instance, in 5G new radio (NR) standard, beam alignment is achieved through \textit{exhaustive beam sweeping}, where both sides search the full codebook to maximize beamforming gain \cite{surveyBA1}.
In practice, the beam sweeping process is executed cyclically to continuously update and maintain the optimal transmit–receive beam pairing. However, this repeated beam sweeping is time-consuming and incurs high latency and energy costs, as each search cycle effectively starts from scratch without leveraging information from previous searches. 
To address these challenges, there is a need for new approaches to ensure efficient and secure  BS–UE beam pair identification in complex and rapidly changing environments. 


{

One of the most promising techniques is the integration of artificial intelligence (AI)-based data-driven approaches for beam alignment. Particularly, deep learning (DL) enables efficient beam alignment by learning non-linear mappings between received signal strength indicators (RSSI) inputs measured over small sized codebook of probing beams and optimal narrow beams, thereby reducing the need for exhaustive beam sweeping. While DL-aided beam alignment techniques achieve notable performance gains, their reliance on correlation-driven mappings and their vulnerability to adversarial inputs remain major challenges \cite{our_work}.   Existing DL-based methods primarily learn statistical correlations from  input RSSI measurements, often ignoring the underlying causal relationships among system variables. Modeling such causal relationships is critical to improve reliability, efficiency, and interpretability of DL models \cite{causal_magazine}. Moreover, the performance of DL-based beam classifiers is highly sensitive to the selected input beams (RSSI measurements), making the identification of a minimal yet informative set of RSSI measurements essential for fast and accurate beam alignment. To this end, recent studies have formulated DL-aided beam alignment as a feature selection problem, where conventional feature selection methods \cite{FastIA} and explainable AI (XAI)–based approaches \cite{KHAN_TWC} are used to identify influential inputs while reducing beam sweeping overhead. However, these approaches do not explicitly account for causal feature relevance and remain vulnerable to adversarial and out-of-distribution inputs, which can significantly degrade beam alignment accuracy and spectral efficiency.

}

To address the limitations of standard AI-based beam alignment approaches, this paper introduces two key contributions: (i) a causal AI–based framework for efficient and reliable mmWave beam alignment, and (ii) a generative defense mechanism based on  variational autoencoders (VAE) that maps perturbed RSSI inputs to clean representations, preserving high prediction accuracy under both normal conditions and adversarial or jamming attacks.


\subsection{Relevant Works}

\textbf{DL-based Beam Alignment:} 
{ DL-aided beam alignment typically requires the BS to transmit reference signals over multiple beams. The UE measures these signals and reports the measurements to the BS, where a neural network selects the optimal serving beam from a predefined codebook. Conventional methods commonly use discrete Fourier transform (DFT)-based quantized beam codebooks, which provide wide angular coverage but limited alignment resolution. Although oversampled discrete Fourier transform (O-DFT) codebooks improve beam granularity, they incur higher training overhead due to the increased number of beams that must be evaluated. To alleviate the signaling overhead,  deep neural networks (DNNs) based solutions have been proposed to accurately identify the optimal
beam by directly sweeping a subset of candidate beams
in the pre-defined codebook \cite{BA4}, explicitly learning probing codebooks along with a beam predictor \cite{BA6}, and DL-based hierarchical search framework to refine best beam search measurements \cite{hierarchical_3}.  Data-driven DL-based solutions have  also been proposed, where  DNNs are designed to exploit auxiliary context, historical channel behavior and prior received signals for beam prediction  \cite{CI1,lim2021deep,ma2021deep}. A major limitation of these prior works \cite{ BA4, BA6, hierarchical_3, CI1,lim2021deep,ma2021deep} on DL-aided
beam alignment is their reliance on explicit propagation
channel knowledge between the BS and all possible UEs
during the training stage, which is difficult to obtain in dynamic environments.  
{Recent studies have shown that the accuracy of DL-based beam alignment strongly depends on both the number and selection of measurement beams \cite{FastIA, KHAN_TWC}. Consequently, input beam selection has been formulated as a feature selection problem, with optimal beam subsets identified using sequential feature selection methods \cite{FastIA} and XAI techniques such as Shapley additive explanations (SHAP) \cite{KHAN_TWC}. However, these approaches typically rely on iterative selection and repeated model retraining, resulting in high computational complexity and limited scalability.}

\textbf{Causal AI for Wireless Communications:}
{Causal AI has recently emerged as a promising paradigm for addressing the limitations of purely data-driven DL approaches in wireless communications \cite{causal_magazine}. While standard DL models predominantly learn statistical dependencies from raw data, causal AI aims to identify high-level variables that encode physically meaningful cause–effect relationships, enabling more robust and interpretable  decisions.}
Causal AI is used to uncover the
cause-effect relationships between the observational data
(state, action, and reward) pertaining to joint communication and sensing environments to facilitate the selection of actions that achieve the highest rewards for both communication and
sensing functionalities \cite{causal1}. Causal AI–based models have been proposed to capture causal relationships among semantic content elements, enabling semantic communication systems to generalize across varying wireless data distributions \cite{causal2, causal3}. Specifically, \cite{causal2} introduces a causal semantic communication framework that learns causally invariant representations from wireless network data. By explicitly modeling the underlying causal structure, the proposed framework improves generalization and robustness under changing network conditions. In contrast, \cite{causal3} investigates causal relationships between explicitly defined semantic variables and downstream task variables by applying the direct linear non-Gaussian acyclic model (DirectLiNGAM) to estimate a causal graph under structural constraints, enabling identification of influential semantic components.
Causal discovery algorithms such as the Peter–Clark  and Greedy Equivalence Search  are utilized for intent-based automation in wireless networks by identifying the causal graph among key performance indications (KPIs), allowing the system to determine which KPIs (e.g., throughput) directly influence a target quality of experience \cite{causal4}.
 {The aforementioned works \cite{causal1, causal2, causal3, causal4} primarily focus on modeling data semantics and uncovering causal relationships among observed system variables to improve generalization and decision making. In contrast, this work proposes a causal AI–based model-agnostic input selection framework for beam alignment, which directly reduces the input dimensionality and beam sweeping overhead. The proposed approach enables fast and accurate beam alignment by limiting the inputs to causally relevant features and offers an efficient alternative to conventional \cite{FastIA} and XAI-based feature selection methods \cite{KHAN_TWC, kHAN-TCOM}, a direction that has not been previously explored.}

\textbf{Robustness against adversarial inputs}: 
Attackers may utilize the broadcast nature of wireless
communication links to generate adversarial attacks on DNN-based classifiers  \cite{attack_beam, my_ICC, AA2}.
In mmWave systems, such attacks can disrupt beam alignment and severely degrade data rates. Limited literature studies have considered the adversarial robustness aspect for mmWave beam prediction in wireless communications \cite{attack_beam, AD2, AD1}. In particular, \cite{attack_beam}  focuses on crafting effective adversarial attacks on beam classification task without providing effective countermeasures against these attacks. 
In BM tasks, adversarial training has shown promise against white-box and black-box attacks \cite{AD2}, but it requires training data for all possible attack scenarios and often overfits to adversarial examples.
Defensive distillation has also been proposed as a mitigation strategy against over-the-air attacks in beam prediction task  \cite{AD1}. However, it is shown to be
vulnerable to high-bounded perturbations. Recently, autoencoders have been proposed for denoising the input data by learning a mapping between potentially perturbed inputs and its corresponding unperturbed input \cite{AD4, AD5}. For instance, denoising autoencoder based defense strategies are proposed for mitigating the effects of adversarial attacks
instantiated on DNN models for power allocation \cite{AD4}, and radio frequency signal classification tasks using autoencoder pre-training \cite{AD5}. While these methods prove effective in mitigating the adversarial impact on input data, the attacks used in these autoencoder based schemes rely on standard fast gradient methods, which generates coarse-grained perturbations with low fooling success rate. Generative models, particularly VAE, can address these issues by learning to denoise a broad spectrum of perturbations, enhancing robustness against practical adversarial/jamming attacks.

\subsection{Contributions}
{In this work, we propose a novel causal AI-based adversarially robust DL beam prediction framework, for the first time in literature.} This framework integrates causal discovery into the BM process to identify the most informative  RSSI inputs for DNN-aided beam alignment in mmWave systems. Crucially, this DNN-based engine operates without explicit channel state information; it trains directly on noisy power measurements from a finite set of DFT codebook-based sensing beams to predict optimal  beams from an O-DFT codebook of narrow beams, ensuring a standards-compliant architecture, and employs a defensive variational autoencoder (DVAE) to enhance robustness against adversarial attacks. The main contributions are summarized as follows:
\begin{itemize}

    \item \textbf{Efficient causal AI-based input selection:}  
    A  novel two-stage {model-agnostic} {causal AI-based }framework is introduced to identify a minimal subset of RSSI inputs for beam prediction. In the former stage, a directed acyclic graph (DAG) is learned to capture dependencies between received powers and the optimal beam.  In the latter, the learned DAG is utilized for causal feature selection. The resulting features closely match those from SHAP-based selection \cite{KHAN_TWC,kHAN-TCOM}, while reducing beam sweeping overhead by 59.4\% and 89.8\% compared to conventional DNN-based selection and exhaustive search, respectively, and achieving comparable accuracy with about $829\times$ lower per-sample computational cost than SHAP.

    \item \textbf{Defense against adversarial attacks:} 
    A lightweight DVAE-based defense mechanism is proposed to mitigate adversarial attacks on the DNN classifier. The DVAE is trained to reconstruct clean RSSI measurements from adversarially corrupted RSSI inputs while preserving performance on clean data. At inference, measurements are first denoised by the DVAE and then passed to the classifier. {The proposed defense outperforms defensive distillation by $\approx15$–$20\%$ under both semi-white and black-box attacks, while sustaining $85\%$ accuracy across different perturbation bounds and recovering the no-attack (clean-input) baseline performance.
}
\end{itemize}

The rest of the paper is organized as follows. Section~\ref{sec:system} presents the system model.  Section~\ref{sec:prob-sol} formulates the beam alignment problem and outlines the proposed solution framework, including the causal AI-based input selection and defense strategy. Section~\ref{CFS_subsection} details the proposed causal feature selection algorithm and the training of the reduced-complexity beam classifier. Section \ref{Attack-BAE} introduces the adversarial attack models and describes the proposed DVAE-based defense strategy. Section \ref{sec:simulation} evaluates the performance of the proposed solution strategy. Finally, conclusions and future research directions are presented in Section \ref{sec:conclusion}.

\begin{figure}
    \centering
{\includegraphics[width=0.4 \linewidth, height=4cm]{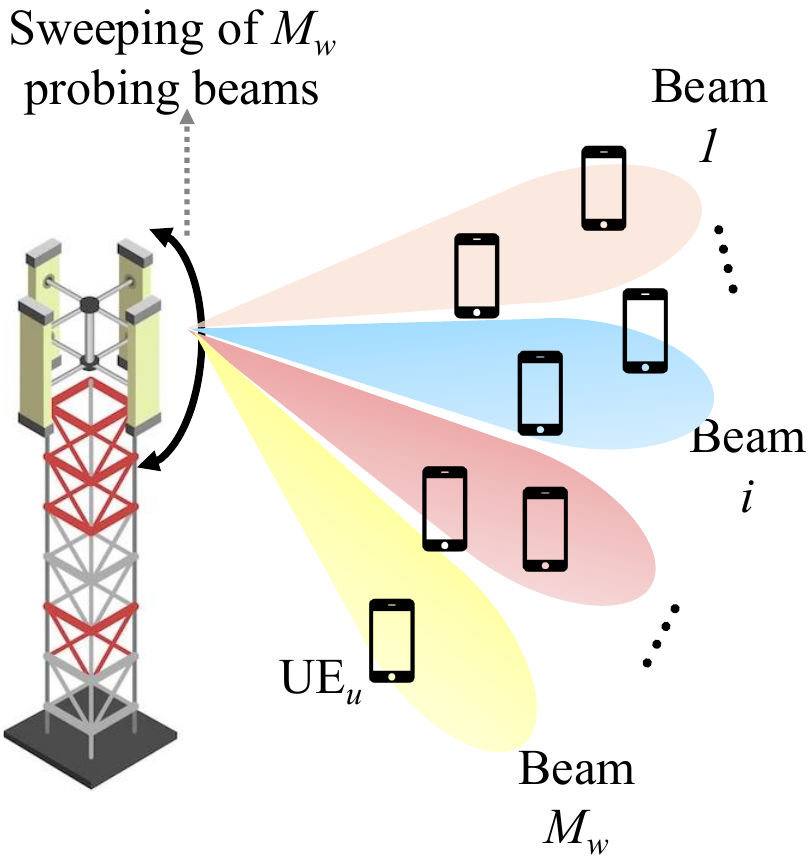}}
      
    \setlength{\belowcaptionskip}{-20pt}
    \caption{System model illustration for the beam alignment in mmWave communication system.}
    \label{fig:sys_model}
\end{figure}

\begin{figure*}
    \centering\includegraphics[width=0.85\linewidth]{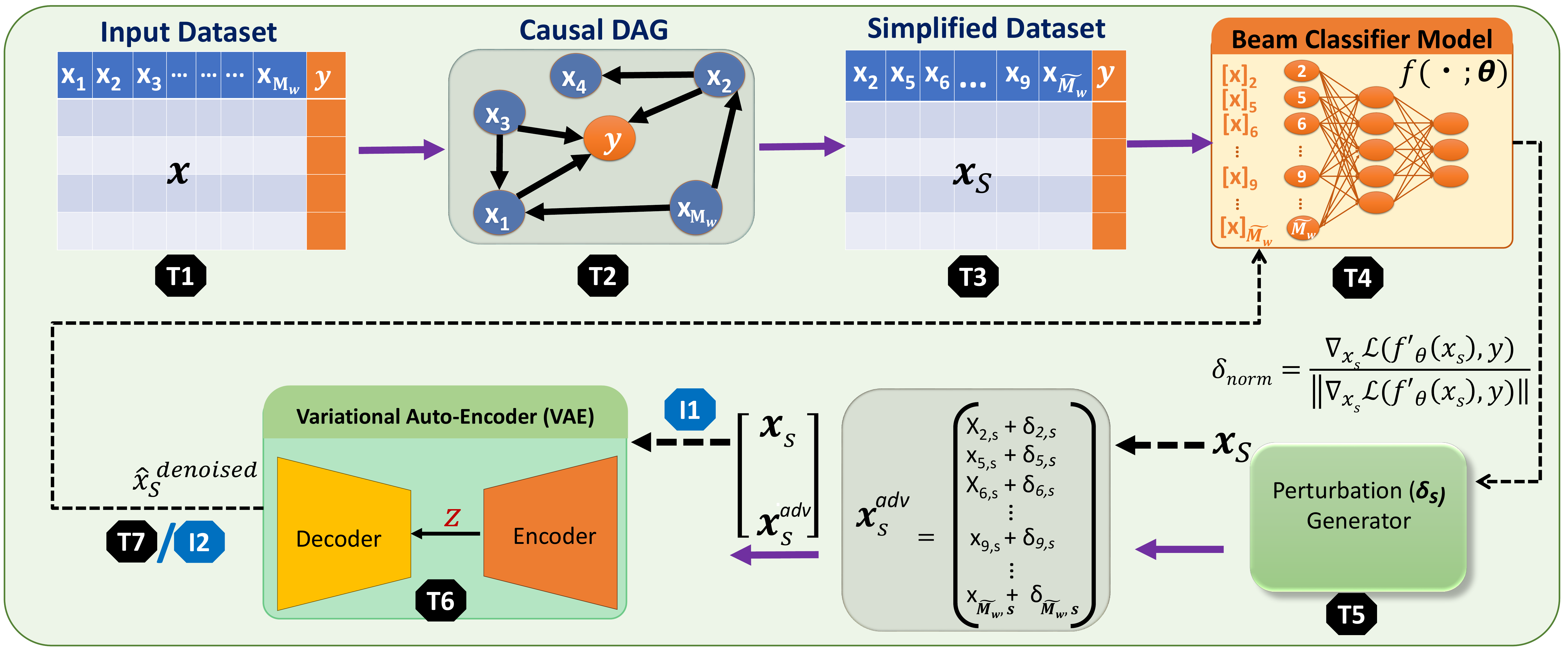}
    \setlength{\belowcaptionskip}{-10pt}
    \caption{Illustration of the proposed causal feature selection and DVAE-assisted robust beam alignment framework, where the BS hosts both the DNN beam classifier and the pre-processing DVAE to mitigate potential adversarial attacks.}
    \label{proposed_flow}
\end{figure*}

\section{System Model} \label{sec:system}
We consider a downlink mmWave communication system, where the BS equipped with a uniform linear array (ULA) with $N_{\mathrm{BS}}$ antenna elements serves $N_\mathrm{U}$ single-antenna
UEs, as illustrated in Fig. \ref{fig:sys_model}. 
For mmWave frequencies, channels often
exhibit sparsity in the angular domain, leading to a constrained number  of multipath components. The  mmWave channel from the BS to the  $\mathrm{UE}_u$  can be described by a multipath geometric channel model as
{\begin{equation}
    \mathbf{h}_u =  \sum_{l=1}^{L}\alpha_{u,l} \mathbf{b}\left(\phi_{u,l}\right),
\end{equation}
where $L$ is the number of paths,}  $\alpha_{u,l}$ represents the complex path gain for the $l$-th path of the $u^{th}$ UE, $\phi_{u,l} $ is the angle of departure for the $l$-th path from the BS to the UE, and $\mathbf{b}\left(\phi_{u,l}^{}\right)$ denotes the beam steering vector, which is given by 
\begin{equation}
    \scalebox{0.96}{$\mathbf{b}(\phi_{u,l}) = \tfrac{1}{\sqrt{N_{\mathrm{BS}}}} \left[1, e^{j \frac{2 \pi}{\lambda} d \sin(\phi_{u,l})}, \dots, e^{j (N_{\mathrm{BS}} - 1) \frac{2 \pi}{\lambda} d \sin(\phi_{u,l})}\right]^T,
    $}
\end{equation}
where $\lambda = \frac{c_0}{f_0}$ denotes the signal wavelength, $c_0$ is the speed of light, $f_0$ refers to the carrier frequency, and $d=\frac{\lambda}{2}$ the antenna spacing. To reduce hardware cost and power consumption, an analog-only beamforming scheme is considered, which uses a single transmit/receive radio frequency chain across $N_{\mathrm{BS}}$ antennas. The BS beamforming vector is thus given by
$
\mathbf{w}=\frac{1}{\sqrt{N_{\mathrm{BS}}}}\left[e^{j \varphi_1}, \ldots, e^{j \varphi_i}, \ldots, e^{j \varphi_{N_{\mathrm{BS}}}}\right]^{\top} \in \mathbb{C}^{N_{\mathrm{BS}} \times 1},
$ 
where $\varphi_i$ denotes the phase shift of the $i$-th antenna element, and the BS employs a beamforming codebook $\mathbf{W} = \{\mathbf{w}_1, \ldots, \mathbf{w}_Y\}$ comprising $Y$ predefined vectors satisfying the constant-modulus constraint $\|\mathbf{w}_y\|^2 = 1$ for $y = 1, \dots, Y$.  During beam sweeping, the BS transmits symbols $s_w \in \mathbb{C}$ to the UEs through the beams in $\mathbf{W} \in \mathbb{C}^{N_{\mathrm{BS}} \times Y}$. The received signal at $\mathrm{UE}_u$ using the $y_u$-th beam $\mathbf{w}_{y_u}$ is
\begin{equation}
r_u = \sqrt{P_{\mathrm{BS}}} \mathbf{h}_u^H \mathbf{w}_{y_u} s_w + z_u,
\end{equation}
where $P_{\mathrm{BS}}$ is the BS transmit power, $\mathbf{h}_u \in \mathbb{C}^{N_{\mathrm{BS}} \times 1}$ is the channel vector, and $z_u$ is additive complex noise with variance $\sigma_z^2$. For unit-power transmitted symbols, the resulting signal-to-noise ratio (SNR) at $\mathrm{UE}_u$ is
\begin{equation}
\text{SNR}_u = \frac{P_{\mathrm{BS}} \left| \mathbf{h}_u^H \mathbf{w}_{y_u} \right|^2}{\sigma_z^2}.
\end{equation}

\section{Problem Formulation and Solution Methodology} \label{sec:prob-sol}
This section describes the problem formulation for the beam
alignment task, introduces our proposed causal input selection
methodology and the DVAE based robustness assessment
framework for beam alignment.
\subsection{Problem Formulation}
{The objective of the beam alignment engine is to optimize the narrow beam delivering the maximum SNR to the $u$-th user. The narrow beams are drawn from an O-DFT codebook of size 
$Y$. For a given BS-UE pair, the optimal beam index $ y_{u}^*$ can be identified as follows:
\begin{align} \label{beam_selection}
    y_{u}^* &= \underset{y_u \in \left\{1,2,\dots,Y\right\}}{\arg \max} 
    \left( \frac{P_{\mathrm{BS}}\left| \mathbf{h}_{u}^H \mathbf{w}_{y_u} \right|^2 }{\sigma_z^2} \right) \notag \\
    &= \underset{y_{u{}} \in \left\{1,2,\dots,Y\right\}}{\arg \max} 
    \left( \left| \mathbf{h}_{u}^H \mathbf{w}_{y_u} \right|^2 \right).
\end{align}}


It is worth noting that identifying $y_{u\mathbf{}}^*$ through exhaustive beam search over $Y$ codewords results in significant beam sweeping overhead.

 
 The considered beam alignment system relies  on RSSI measurements from a small set of wide sensing beams $M_{\mathrm{w}}$ ($M_{\mathrm{w}} \ll Y$) to select the optimal narrow beam from an O-DFT codebook of size $Y$, thereby avoiding exhaustive search. During beam sweeping, the BS transmits pilots sequentially over the $M_{\mathrm{w}}$ beams, and all UEs measure and report the corresponding RSSI values. It is assumed that beam sweeping, measurement, and reporting occur within the coherence time during which the channel remains constant. The reported beam sweeping results for  UE  $u$ are given by
\begin{equation}\label{rp}
\mathbf{x}_u={\left[x_{u_1},\ldots,x_{u_{M_\mathrm{w}}} \right]}=\left[\left|r_{u_1}\right|^2 \cdots\left|r_{u_{M_{\mathrm{w}}}}\right|^2\right]^T ,
\end{equation}
{where $r_{u_j}$ is the received signal using the $j$-th sensing beam, $\forall j \in M_{\mathrm{w}}$}. Notably, the overall number of beams swept remains constant,
unaffected by the number of UEs, as multiple UEs can
measure the sensing beams concurrently, utilizing shared time
and frequency resources.

{While DL-based methods can map RSSI measurements from $M_{\mathrm{w}}$ sensing beams to the optimal narrow beam index, their accuracy strongly depends on the choice and number of sensing beams.
Increasing the number of sensing beams reduces prediction error but incurs higher latency. Consequently, an essential trade-off exists between beam acquisition latency and the quality of the selected sensing beams. Further, the underlying
DL remains vulnerable to adversarial inputs.}


}


\subsection{Solution Methodology}

To mitigate high beam-sweeping overhead and adversarial vulnerability in existing DL-based beam alignment methods, we propose a framework in which the BS employs a smart feature selection strategy to extract causally relevant inputs for beam classification, and integrates a DVAE module that maps perturbed inputs back to their clean versions for adversarial robustness. {The proposed framework involves two stages (i) an offline feature selection and training stage (ii) an online inference stage.}  As illustrated in Fig. \ref{proposed_flow} and introduced in the sequel sections, the system operation proceeds as follows:
\subsubsection*{\textbf{Training Stage:}}
\begin{itemize}
    \item[\Large \textcircled{\footnotesize{T1}}] Site-specific labeled dataset  is first generated with the RSSI values as the input features and corresponding best beam index as ground-truth label, [c.f. Section \ref{DA_subsection}].    
    \item[\Large \textcircled{\footnotesize{T2}}] Causal discovery is employed to identify the underlying cause–effect relationships between feature attributes and the target variable from observational data, using a causal directed acyclic graph (DAG), [c.f. Section \ref{CBS}]. 
    \item[\Large \textcircled{\footnotesize{T3}}] The causally relevant features are extracted from the constructed DAG  using a novel causal feature selection strategy to obtain a simplified dataset having a subset of model inputs, i.e, RSSI values, that significantly contribute to the best beam prediction,  [c.f. Section \ref{CBS}].
    \item[\Large \textcircled{\footnotesize{T4}}] A simplified DL model is trained on the simplified dataset extracted via causal discovery to predict the optimal beam directions [c.f. Section \ref{Simplified_DNN_training}]. 
    \item [\Large \textcircled{\footnotesize{T5}}] Adversarial perturbations are generated using the simplified  classifier's gradients  for launching semi-white box attack as well as training dedicated surrogate model for crafting black-box attacks [c.f. Section \ref{BAE-Attack}].
     \item [\Large \textcircled{\footnotesize{T6}}] 
     The proposed defense methodology trains the DVAE  to map perturbed adversarial samples to their corresponding clean (non-adversarial) representations. Specifically, the DVAE is trained on combinations of original clean inputs and their adversarially perturbed representation to denoise adversarial inputs  and retain performance on non-adversarial samples [c.f. Section \ref{Attack-BAE-Defense}].   
     \item [\Large \textcircled{\footnotesize{T7}}]
     {To further enhance beam alignment performance in adversarial settings, the reconstructed representations, i.e, the DVAE outputs,  are used to retrain and fine-tune the downstream beam classifier in an end-to-end manner prior to online deployment [c.f. Section \ref{fine-tune-BAE}]. }
     
\end{itemize}
\subsubsection*{\textbf{Inference Stage:}}
\begin{itemize}
   { \item [\Large \textcircled{\footnotesize{I1}}]
     During inference, the received RSSI inputs are first pre-processed using the DVAE, i.e, incoming beam measurements are first forward propagated through the DVAE to filter any potentially added adversarial noise before being inputted to the simplified DNN.
      \item [\Large \textcircled{\footnotesize{I2}}]
    The denoised outputs of the DVAE are used as inputs to the simplified DNN beam classifier.}
    \end{itemize}

\section{Causality-based Beam Selection}\label{CFS_subsection}
 In this section, we treat the DL model input selection as a causal feature selection problem, aiming to identify inputs that have a true causal influence on the target. This is achieved by modeling the collected data as a DAG structure, where each node in the constructed DAG represents either an input or the target beam index, and edges reflect potential causal dependencies. The goal is to isolate the subset of RSSI feature nodes, i.e, sensing beam measurements, that have a direct causal influence on selected best narrow beam, effectively filtering out redundant or spurious features.
 
 
\subsection{Data Acquisition}\label{DA_subsection}
To reduce signaling overhead caused by sweeping the entire set of sensing beams, we leverage the causal connections between the RSSI measurements  and the {optimal
beam index (target variable)} to identify a subset of input RSSIs that contribute more significantly to the beam selection process. This approach requires collecting the data from the site-specific environment as a first step. We propose collecting
RSSI measurements at each UE during a data collection phase. The ground-truth label $y_u$ for each data sample $\mathbf{x}_u$ is obtained  by searching the optimal beam direction in the O-DFT codebook by performing noise-free exhaustive search. The constructed  dataset is denoted by \( \mathcal{D} = \{ (\mathbf{x}_u, y_u) \}_{u=1}^{N} \), where \( \mathbf{x}_u \in \mathbb{R}^{M_{\mathrm{w}}} \) represents the beam sweeping measurements for user \( u \),  \( y_u \in \{1, 2, \ldots, Y\} \) denotes the ground-truth label corresponding to the optimal beam index, generated using \eqref{beam_selection}, and  \( N \) denotes the total number of samples in the dataset.

\subsection{Causal Beam Discovery} \label{CBS}
We model the  causal relationships between the input features and the  target variable using a Bayesian graph \( \mathcal{G} = \{\mathcal {V}, \mathcal{E}\} \), where each node in \( \mathcal{V} \) corresponds to a feature in the dataset, including the RSSI values \( \mathbf{x}_u \in \mathbb{R}^{M_{\mathrm{w}}} \) and the target beam index \( y_u \in \{1, \ldots, Y\} \). The directed edges \( \mathcal{E}\) represent causal influences among these features. The goal of causal discovery is to estimate the causal structure  from observed samples \( \{ (\mathbf{x}_u, y_u) \}_{u=1}^{N} \). We assume that the target \( y_u \) is not a cause of any RSSI feature, i.e., \( (y_u, x_{u_i}) \notin \mathcal{E} \); instead, $y_u$ is treated as an effect, in line with the unidirectional nature of the underlying physical process. \footnote{{This assumption is justified as the optimal beam index is determined by the underlying propagation environment. Encoding this domain knowledge as a structural constraint in the DAG anchors the causal interpretation and prevents learning spurious reverse edges.}} To estimate the causal  DAG structure, we use the direct method for learning a linear non-Gaussian structural equation model (DirectLiNGAM) algorithm \cite{directlingam}. The method iteratively applies least squares regressions to compute residuals and selects the variable most independent from its residuals to build a causal ordering. This ordering is then used to estimate connection strengths between features. The resulting DAG encodes directional dependencies among the dataset's features and the target variable.

\begin{figure}[t]
    \centering
    \includegraphics[width=1\linewidth]{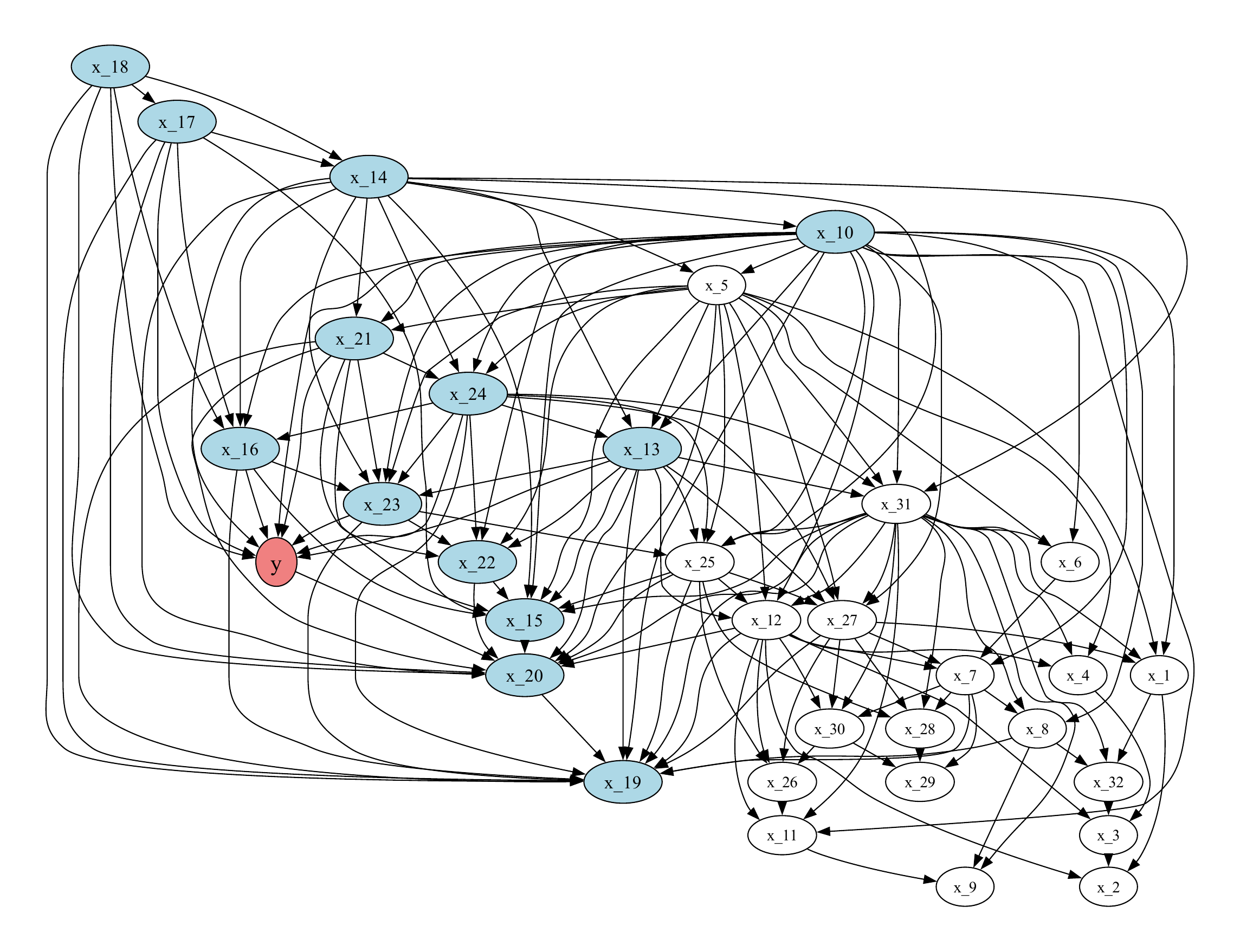}
  \setlength{\abovecaptionskip}{-5pt}
    \setlength{\belowcaptionskip}{-10pt}
    \caption{DAG obtained via DirectLiNGAM based causal beam selection approach.}
    \label{fig:DAG}
\end{figure}

As shown in Fig.~\ref{fig:DAG}, the output of DirectLiNGAM is a causal graph where each node represents a dataset feature, and a directed edge \( x_i \xrightarrow{} x_j \) indicates that feature \( x_i \) causes feature \( x_j \).
The DirectLiNGAM algorithm estimates the underlying causal structure among input features and the target variable, assuming a linear non-Gaussian data-generating process. Given the dataset \( \mathcal{D} = \{ (\mathbf{x}_u, y_u) \}_{u=1}^N \), we define the joint variable 
$
\mathbf{z} = [x_{u_1}, \dots, x_{u_{M_{\mathrm{w}}}}, y_u]^\top \in \mathbb{R}^p,
$ 
where $p = M_{\mathrm{w}} + 1$. The causal generative model is formulated as  \cite{directlingam}
\begin{equation}
\mathbf{z} = \mathbf{B}\mathbf{z} + \mathbf{e},
\end{equation}
{where $\mathbf{B} \in \mathbb{R}^{p \times p}$ is a strictly lower-triangular matrix encoding directed causal effects, and $\mathbf{e} \in \mathbb{R}^p$ is a vector of mutually independent non-Gaussian disturbances. {The non-Gaussianity of $\mathbf{e}$ is the key property enabling  causal {direction} to be identified from observational data alone, allowing DirectLiNGAM to recover directed edges   \cite{directlingam}. } A nonzero element $b_{ij} \neq 0$ implies a directed edge $z_j \rightarrow z_i$ in the underlying DAG. To identify the causal ordering, each variable $z_j$ is sequentially regressed onto all other variables $z_i$ ($i \neq j$), yielding residuals $r_i^{(j)} = z_i - \hat{\beta}_{ij} z_j$, where $\hat{\beta}_{ij}$ is the least-squares regression coefficient. The exogenous variable is then detected as the one most independent from its residuals. This procedure is applied recursively to the remaining variables until a causal ordering is obtained. Finally, the direct causal effects (entries of $\mathbf{B}$) are estimated via least squares following the determined ordering.

{The proposed causal beam selection approach is summarized in Algorithm~\ref{alg:CFS}.  The inputs of the algorithm are the dataset \( \mathcal{D}\) and maximum feature count \( \widetilde{M}_{\mathrm{w}}  \). The output is a subset of causally relevant RSSI measurements \( \mathbf{x}_{u,S} = \{x_{u_i} \mid i \in \mathcal{S} \} \), where \( \widetilde{M}_{\mathrm{w}} = |\mathcal{S}| \ll M_{\mathrm{w}} \).}  The effect matrix $\mathbf{B}$ is first estimated using the DirectLiNGAM procedure, which iteratively regresses each variable in $\mathbf{z}$ onto others and ranks them based on residual independence, resulting in a causal ordering from which direct causal effects are computed.} Next, the direct parents of the target variable $y$, {denoted by $\mathcal{P}(y)$},  are determined from the estimated effect matrix $\mathbf{B}$ as the subset of feature variables for which the corresponding entries $B_{y,i} \neq 0$ ($i \leq M_{\mathrm{w}}$). {Each nonzero coefficient $B_{y,i}$ quantifies the magnitude of the direct causal effect from the feature $x_{u_i}$ to the target $y$, indicating that $x_{u_i}$ serves as a direct parent of $y$ in the learned causal graph (Lines 3--4). }
 To ensure inclusion of other potentially relevant variables, the algorithm computes the total connectivity for each feature \( i \) (Lines 5--9), defined as the sum of its incoming connections (number of nonzero effects from other features to \( i \)) and outgoing connections (number of nonzero effects from \( i \) to other features), and ranks features accordingly (Line 11). Finally, the selected feature set \( \mathcal{S} \) is constructed by combining the direct parents of the target with top-ranked features based on total connectivity until \( |\mathcal{S}| = \widetilde{M}_{\mathrm{w}}  \) (Lines 13--18). 


 {\textbf{Remark:} The proposed method operates under the \emph{observational causal 
discovery} paradigm and does not consider interventional causality in the 
do-calculus sense~\cite{pearl2016causal}. Rather, under LiNGAM identifiability conditions, the DirectLiNGAM procedure recovers the Markov equivalence class and a unique causal ordering consistent with the data-generating process, providing a strictly stronger guarantee than SHAP, which explains model outputs based on marginal feature contributions rather than causal structure.

}

\subsection{Training of a Simplified DNN-based Classifier} \label{Simplified_DNN_training}
 {The subset of features obtained using Algorithm~\ref{alg:CFS} is used to train a simplified DNN-based classifier with input size \(\widetilde{M}_{\mathrm{w}}\), thereby reducing both sensing and inference complexity relative to the full DNN model that utilizes RSSI measurements from all \(M_{\mathrm{w}}\) sensing beams. 
 

{As illustrated in Fig.~\ref{proposed_flow}, the simplified DNN deployed at the BS is denoted by 
$f(\hspace{0.05cm}\cdot\hspace{0.05cm};\boldsymbol{\theta}): \mathbb{R}^{\widetilde{M}_{\mathrm{w}}} \rightarrow \mathbb{R}^{Y}$, parameterized by $\boldsymbol{\theta}$. The input to the DNN is the reduced RSSI feature vector $\mathbf{x}_{u,S}\in\mathbb{R}^{\widetilde{M}_{\mathrm{w}}}$, obtained by retaining only the $\widetilde{M}_{\mathrm{w}}$ causally relevant RSSI measurements from the original dataset $\mathcal{D}$ using the method described in Section~\ref{CBS}, while the output is a $Y$-dimensional score vector corresponding to the $Y$ candidate narrow beams, with the predicted beam given by the index of the maximum score. To train the simplified classifier for narrow-beam selection, the BS collects RSSI measurements only from the $\widetilde{M}_{\mathrm{w}}$ selected sensing beams, yielding the reduced training dataset $\widetilde{\mathcal{D}}={(\mathbf{x}_{u,S}, y_u)}_{u=1}^{N}$, where $N$ is the number of samples.
Given an input $\mathbf{x}_{u,S}$, the classifier produces a beam score vector $f'(\mathbf{x}_{u,S};\boldsymbol{\theta}) \in \mathbb{R}^{Y}$, 
where each element represents the predicted score (e.g., softmax output) of a candidate beam. 
The top-$K$ beams can be selected by ranking these scores in descending order as 
$\{\hat{i}_1,\hat{i}_2,\ldots,\hat{i}_K\}=\text{argsort}(f'(\mathbf{x}_{u,S};\boldsymbol{\theta}))$, 
where $\text{argsort}(\cdot)$ returns the indices of the $K$ highest-scoring beams. 
The network parameters $\boldsymbol{\theta}$ are optimized using the  cross-entropy loss between the predicted beam score  and the true beam label, defined as
}
\begin{equation}\label{eq:ce_loss}
\mathcal{L}_{\mathrm{CE}}(\boldsymbol{\theta}; \mathcal{\widetilde{D}}) 
= -\frac{1}{N} \sum_{u=1}^{N} \sum_{y=1}^{Y} p_{u,y}^{\star} 
\log \widehat{p}_{u,y},
\end{equation}
where $p_{u,y}^{\star}$ denotes the true label distribution such that $p_{u,y}^{\star}=1$ if the $y_u$-th candidate narrow beam is the actual
optimal beam and $0$ otherwise, and $\widehat{p}_{u,y} \in (0,1)$ represents the predicted probability of selecting beam $y$, obtained by applying the softmax function to the DNN output logits.}

{From a practical perspective, the proposed causal beam selection framework operates in offline training and online deployment phases. Training can be performed using site-specific digital twins and ray-tracing simulations prior to deployment. By identifying the most informative  causal sensing beams that capture dominant propagation characteristics, the framework reduces sensing and inference complexity while maintaining beam alignment performance. Since the causal relationships depend on the propagation environment, significant environmental changes may require feature reselection and DNN retraining/fine-tuning,  although such updates are expected to be infrequent in practice due to slow evolution of large-scale propagation characteristics}

{Next, we discuss the generation of adversarial attacks on the simplified DNN $f(\hspace{0.05cm}\cdot\hspace{0.05cm};\boldsymbol{\theta})$ and our methodology for defending the simplified beam classifier against such attacks. Crafting these attacks requires the gradient of the loss function with respect to the model; thus, the adversary may either use gradient information from the simplified DNN deployed at the BS or rely on a surrogate network, corresponding to semi-white box and black-box threat models, respectively.
}

\begin{algorithm}[t]
\caption{Causal Beam Selection Algorithm}
\label{alg:CFS}
\footnotesize
\SetKwInOut{Input}{Input}
\SetKwInOut{Output}{Output}
\Input{Dataset \( \mathcal{D} \), maximum feature count \( \widetilde{M}_{\mathrm{w}} \)}
\Output{Subset of input features \( \mathbf{x}_S \)}
Train DirectLiNGAM on \( \mathcal{D} \) to obtain  matrix \( \mathbf{B} \)\;
Derive causal ordering from residual independence\;
Select direct parents of \( y \)\;
\quad \( \mathcal{P}(y) \gets \{ i \mid B_{y,i} \neq 0 \} \)\;

\textbf{Compute total connectivity for each feature \( i \)}\;
\For{\( i \gets 1 \) \KwTo \( p \)}{
    \( \text{In}(i) \gets \sum_j \mathbf{1}_{B_{i,j} \neq 0} \)\;
    \( \text{Out}(i) \gets \sum_j \mathbf{1}_{B_{j,i} \neq 0} \)\;
    \( \text{Conn}(i) \gets \text{In}(i) + \text{Out}(i) \)\;
}
Sort features by \( \text{Conn}(i) \) in descending order\;
Initialize \( \mathcal{S} \gets \mathcal{P}(y) \)\;

\While{\( |\mathcal{S}| < \widetilde{M}_{\mathrm{w}}  \)}{
    Add next top-ranked feature by total connectivity to \( \mathcal{S} \)\;
}
Trim \( \mathcal{S} \) to size \( \widetilde{M}_{\mathrm{w}} \): \( \mathcal{S} \gets \mathcal{S}[:\widetilde{M}_{\mathrm{w}} ] \)\;
Input subset: \( \mathbf{x}_S \gets \{x_i \mid i \in \mathcal{S} \} \)\;
\Return \( \mathbf{x}_S \)\;
\end{algorithm}

\section{Adversarial Attacks and Defense \\ for Beam Prediction Model} \label{Attack-BAE}
In this section, we formulate adversarial attacks on the simplified DNN-based beam classifier under different threat models. We consider semi-white box attacks with full model knowledge and black-box attacks that rely on surrogate models to craft over-the-air attacks. We also present a VAE-based defense mechanism for mitigating adversarial attacks instantiated on the simplified DNN beam classifier.

\subsection{Attack Formulation}\label{BAE-Attack}
Although causal feature selection reduces input dimensionality and signaling overhead, the simplified DNN remains vulnerable to adversarial perturbations, where a jammer/adversary can manipulate RSSI inputs by introducing carefully crafted perturbations to the input RSSI values to induce incorrect beam predictions. Notably, adversarial examples exhibit a transferability property \cite{yuan2019adversarial}, whereby attacks designed to mislead one model can often induce erroneous outputs in other models trained for the same task. As a result, an adversary does not require full knowledge of the simplified DNN’s architecture or parameters  to launch a successful attack. Instead, attacks can be generated using surrogate models and subsequently deployed against the target DNN to induce beam misclassifications.

We consider two attack models depending on the knowledge available to the adversary for crafting the attack:

\begin{enumerate}
\item \textit{Semi-white box attack}, where the adversary has full knowledge of the simplified DNN at the BS and its parameters but is blind to the defense, 
e.g., denoising performed on the measurements before inference. { In contrast, a white-box adversary would also know the DVAE-based defense and its architecture, along with full knowledge of the simplified DNN at the BS, and all associated parameters.} Thus, under the semi–white box threat model, the adversary can still exploit the model’s gradients to craft perturbations that directly target the deployed DNN-based classifier. 

\item \textit{Black box attack}, which  relies on the transferability property of adversarial attacks, as the adversary lacks direct access to the simplified DNN at the BS. Since the model architecture is unknown
to the attacker, a surrogate model which emulates the original
model has to be trained. Due to the lack of direct model access, such attacks are generally less potent than their semi–white box counterparts but are more pragmatic in real-world environments as
adversary will rarely have precise information about the simplified DNN architecture and its parameters.

\end{enumerate}

In our setting, the BS receives RSSI measurements using \(\widetilde{M}_{\mathrm{w}}\) selected sensing beams. Meanwhile, the adversary generates an adversarial attack by injecting an adversarial perturbation \(\boldsymbol{\delta}_{u,S}=[\delta_{u_1},\ldots,\delta_{u_{\widetilde{M}_{\mathrm{w}}}}]^{\mathsf{T}}\in\mathbb{R}^{\widetilde{M}_{\mathrm{w}}}\) to manipulate the input, yielding
\begin{equation}
\mathbf{x}_{u,S}^{\mathrm{adv}}=\mathbf{x}_{u,S}+\boldsymbol{\delta}_{u,S},
\label{eq:adv_input}
\end{equation}
where \(\mathbf{x}_{u,S}^{\mathrm{adv}}\) represents the perturbed adversarial sample and \(\mathbf{x}_{u,S}\) is the clean input vector of RSSI values over \(\widetilde{M}_{\mathrm{w}}\) beams for the \(u\)-th sample. The goal of the adversary is to degrade the beam selection accuracy while keeping the perturbation imperceptible, which can be formulated as the following constrained optimization problem:
\begin{subequations}\label{eq:optimization_problem}
\begin{equation}
\min_{\boldsymbol{\delta}_{u,S}} \ \|\boldsymbol{\delta}_{u,S}\|_{2}
\end{equation}
\begin{equation}\label{eq:misclassify_constraint}
\text{s.t.} \quad 
{\arg\max_{j} f'_{j}\big(\mathbf{x}_{u,S}^{\mathrm{adv}};\boldsymbol{\theta}\big) \neq \arg\max_{j} f'_{j}\big(\mathbf{x}_{u,S};\boldsymbol{\theta}\big)}
\end{equation}
\begin{equation}\label{eq:input_space_constraint}
\mathbf{x}_{\min} \le \mathbf{x}_{u,S}^{\mathrm{adv}} \le \mathbf{x}_{\max}
\end{equation}
\end{subequations}

Here,  \(\|\cdot\|_{2}\) denotes the \(\ell_{2}\)-norm and quantifies the perturbation power added to the reported RSSI values, and {$ f'_{j}\big(\hspace{0.01cm}\cdot\hspace{0.01cm};\boldsymbol{\theta}\big) $ is the DNN classifier output corresponding to the j-th beam}. The adversary’s goal is to design \(\boldsymbol{\delta}_{u,S}\) that forces the DNN to misclassify the predicted beam, as in \eqref{eq:misclassify_constraint}, while ensuring that the perturbed sample \(\mathbf{x}_{u,S}^{\mathrm{adv}}\) remains within the range of valid received power values \([\mathbf{x}_{\min},\mathbf{x}_{\max}]\). {Due to the nonlinearity of the DNN, \eqref{eq:optimization_problem} is generally intractable in closed form and is typically approximated using gradient-based  methods \cite{AD4}.}

\subsection{Adversarial Data Generation}
In the \textbf{semi-white box attack model}, the adversary is assumed to have full access to the simplified DNN model {$f(\mathbf{x}_{u,S};\boldsymbol{\theta})$}, including its architecture and parameters. {The attacker therefore computes the gradient of the beam-classification loss with respect to the input, expressed as $\mathbf{g}_u = \nabla_{\mathbf{x}_{u,S}} \mathcal{L}\big(f(\mathbf{x}_{u,S};\boldsymbol{\theta})\big)$, and generates the perturbation by normalizing and scaling this gradient as $\boldsymbol{\delta}_{u,S} = \epsilon \frac{\mathbf{g}_u}{\|\mathbf{g}_u\|_2}$, where $\epsilon$ denotes the perturbation budget. The adversarial input is then formed by adding the adversarial perturbation to the unperturbed inputs  according to (\ref{eq:adv_input}).} This attack model reflects scenarios in which the DNN deployed at the BS is publicly available or based on standardized architectures, providing the adversary complete model knowledge.

In the \textbf{black box attack model}, the adversary is aware of the beam selection task at the BS but lacks access to the simplified DNN's architecture, and parameter values.{ Here, the adversary exploits the transferability of adversarial attacks by training a surrogate model and using its gradients to craft perturbations, which are then applied to the inputs used by the simplified DNN at the BS. To mount an attack, the adversary trains a surrogate neural network $\widetilde{f}(\mathbf{x}_{u,S};\boldsymbol{\widetilde{\theta}})$, with weight parameters $\boldsymbol{\widetilde{\theta}}$, for the same beam selection task by first obtaining clean samples used to train \(f'(\mathbf{x}_{u,S};\boldsymbol{\theta})\). The attacker then computes the surrogate model's gradient $\widetilde{\mathbf{g}}_u = \nabla_{\mathbf{x}_{u,S}} \mathcal{L}\big(\widetilde{f}(\mathbf{x}_{u,S};\boldsymbol{\widetilde{\theta}})\big)$ and crafts the perturbation as $\boldsymbol{\delta}_{u,S} = \epsilon \frac{\widetilde{\mathbf{g}}_u}{\|\widetilde{\mathbf{g}}_u\|_2}$, where $\widetilde{\mathbf{g}}_u$ denotes the loss gradient of surrogate network with respect to the input. The crafted perturbations are then injected into input samples according to (\ref{eq:adv_input}), before they are processed by the simplified DNN classifier at the BS. Although the gradient used to generate such attacks points in a suboptimal direction i.e., different from the gradient loss calculated using the DNN under attack at the BS, these perturbations can still effectively degrade the beam alignment performance.} 

Next, we present a VAE-based strategy, which reconstructs clean RSSI from corrupted inputs while mapping benign RSSI to near-identical outputs without performance degradation.
\subsection{VAE-based Defense}\label{Attack-BAE-Defense}
This section presents our methodology for defending the  DNN beam classifier against adversarial attacks.  {As illustrated in \Large \textcircled{\footnotesize{T6}} \normalsize in Fig. \ref{proposed_flow}},
we introduce a DVAE module to pre-process inputs before they are fed into the DNN.  During inference, adversarial inputs are first denoised by the trained VAE and subsequently provided to the DNN for beam selection. 

The VAE-based defense combines the autoencoder architecture with a variational Bayesian approach for statistical inference \cite{creswell2018denoising}. In our framework, the VAE is designed to reconstruct clean RSSI vectors $\mathbf{x}_{u,S}$ from potentially corrupted observations $\mathbf{x}_{u,S}^{\mathrm{adv}}$ using a probabilistic mapping. Specifically, the encoder model $\mathscr{E}_{\boldsymbol{\Phi}}$ transforms the input $\tilde{\mathbf{x}}_{u,S}$, where $\tilde{\mathbf{x}}_{u,S} \in \{\mathbf{x}_{u,S}, \mathbf{x}_{u,S}^{\mathrm{adv}}\}$, into a latent distribution parameterized by mean and variance 
$\boldsymbol{\mu}_{z_u}, \boldsymbol{\sigma}_{z_u}^2 = \mathscr{E}_{\boldsymbol{\Phi}}(\tilde{\mathbf{x}}_{u,S}).$ 
A latent sample $\mathbf{z}_u$ is then drawn using the reparameterization 
$\mathbf{z}_u = \boldsymbol{\mu}_{z_u} + \boldsymbol{\sigma}_{z_u} \odot  \boldsymbol{\epsilon'}, \quad  \boldsymbol{\epsilon'} \sim \mathcal{N}(\mathbf{0}, \mathbf{I}),$ where $\odot$ is the Hadamard product and $\mathbf{z}_u$ obeys the normal distribution $\mathcal{N}(\boldsymbol{0}, \boldsymbol{\sigma}_{z_u}^{2})$ with $\mathbf{0}$ denoting the all-zeros vector.
Reparameterization
 allows backpropagation through the stochastic sampling while capturing the variability of the latent space.
 The decoder $\mathscr{D}_{\boldsymbol{\Theta}}$ maps $\mathbf{z}_u$ to an approximate reconstruction of the original data sample $\hat{\mathbf{x}}_{u,S}$, such that $
    \hat{\mathbf{x}}_{u,S} = \mathscr{D}_{\boldsymbol{\Theta}}\left(\mathscr{E}_{\boldsymbol{\Phi}}(\tilde{\mathbf{x}}_{u,S})\right),
$ where, $\mathscr{E}_{\boldsymbol{\Phi}}$ and $\mathscr{D}_{\boldsymbol{\Theta}}$ denote the encoder and decoder with parameters $\boldsymbol{\Phi}$ and $\boldsymbol{\Theta}$, respectively. By regularizing the latent space toward a standard Gaussian prior,
the VAE preserves the essential structure of clean RSSI representations while suppressing adversarial perturbations.

The complete procedure is summarized in Algorithm~\ref{alg:adv_vae_denoising}. 
The algorithm takes as inputs the clean RSSI samples $\{\mathbf{x}_{u,S}\}_{u=1}^N$, 
the trained beam selection DNN $f(\cdot;\boldsymbol{\theta})$, a perturbation bound $\epsilon$, 
and the randomly initialized encoder $\mathscr{E}_{\boldsymbol{\Phi}}$ and decoder $\mathscr{D}_{\boldsymbol{\Theta}}$ functions. 
The output is the beam index $\hat{y}$ obtained by passing the denoised inputs $\hat{\mathbf{x}}_{\text{denoised}}$ 
to the beam classifier. First, for each input RSSI sample $u$, the algorithm computes the gradient $\mathbf{g}_u$ of the beam 
classification loss with respect to the input. The adversarial perturbation is then generated by scaling the normalized gradient $\boldsymbol{\delta_{\text{norm}}}=\frac{\mathbf{g}_u}{\|\mathbf{g}_u\|_2},$ with $\epsilon$, 
and the resulting adversarial input is formed by adding the perturbation to the clean input (Lines 3--5). During training, concatenated clean and adversarial RSSI samples form the input
set $\mathbf{X}_{\text{in}}$, while the corresponding clean RSSI samples form
the target set $\mathbf{X}_{\text{target}}$ (Lines 7--9). 
This ensures that the VAE learns to denoise adversarial inputs while reconstructing clean inputs, thereby preserving performance in both attack and non-attack scenarios. 
For each minibatch, the VAE produces a reconstruction $\hat{\mathbf{x}}_{u,S}$ of the input sample $\mathbf{x}_{u,S}$.
The training loss is defined as a weighted sum of the reconstruction error and 
the Kullback--Leibler (KL) divergence:  
\begin{equation}\label{loss_dVAE}
\begin{split}
\mathcal{L}_{\text{DVAE}} &=\!\!\!\!\!\!
\underbrace{\ell_r(\mathbf{x}_{u,S}, \hat{\mathbf{x}}_{u,S})}_{\text{Reconstruction (denoising)}} 
+
\underbrace{\ell_{D_\text{KL}}\big(\mathcal{N}(\boldsymbol{\mu}_{z_u}, \boldsymbol{\sigma}_{z_u}^2)\,\|\,\mathcal{N}(\mathbf{0}, \mathbf{I})\big)}_{\text{Latent regularization}}\\&= 
\tfrac{1}{N} \sum_{u=1}^{N} \Bigg[
\alpha \, \big\| \hat{\mathbf{x}}_{u,S} - \mathbf{x}_{u,S} \big\|_2^2 \\
&\qquad \!\!+\!\! \; \beta \,
\tfrac{1}{2} \sum_{k=1}^{K} 
\big( \mu_{z_u,k}^2 \!+\! \sigma_{z_u,k}^2 \!\!-\!\! \log \sigma_{z_u,k}^2 - 1 \big)
\Bigg],
\end{split}
\end{equation}
where the first term $\ell_r(\mathbf{x}_{u,S}, \hat{\mathbf{x}}_{u,S})$ measures the squared 
reconstruction error between the VAE’s reconstruction of an adversarial
example and its corresponding clean sample, while the second term 
$\ell_{D_\text{KL}}$ 
regularizes the latent space by computing the KL divergence between the approximate posterior 
$\mathcal{N}(\boldsymbol{\mu}_{z_u}, \boldsymbol{\sigma}_{z_u}^2)$ and the standard normal prior 
$\mathcal{N}(\mathbf{0}, \mathbf{I})$ across the $K$ latent dimensions. 
The hyperparameters $\alpha$ and $\beta$ control the trade-off, with $\alpha \gg \beta$ 
to prioritize reconstruction accuracy. The VAE parameters $(\boldsymbol{\Phi}, \boldsymbol{\Theta})$ are updated 
via gradient descent on the loss (\ref{loss_dVAE}) (Lines 10--14). {As shown in \Large \textcircled{\footnotesize{I2}} \normalsize in Fig. \ref{proposed_flow}}, during inference stage, each received test input $\mathbf{x}_{u,S}^{\text{test}}$, whether adversarially perturbed or clean, is first processed 
through the trained VAE to produce a denoised RSSI vector $\hat{\mathbf{x}}_{u,S}^{\text{denoised}}$ (Line 17). 
This denoised signal is then passed into the beam selection DNN to predict the final beam index $\hat{y}$ (Line 18). 
By training the VAE to map both clean and perturbed inputs to clean representations, the system mitigates adversarial perturbations while preserving performance on benign inputs.

Note that in the implementation of Algorithm~\ref{alg:adv_vae_denoising}, adversarial perturbations (Line 3) are crafted using the gradient information of the simplified DNN deployed at the BS, and added to the legitimate inputs as described in \eqref{eq:adv_input}. This approach corresponds to a semi-white box attack, which is selected due to its higher potency compared to black box attacks. {Consequently, training the system to defend against semi-white box perturbations is expected to enhance robustness against black box attacks as the  semi-white box attacks directly exploit the gradient information of BS classifier, exposing the model to transferable adversarial patterns commonly produced by surrogate, black box attackers.}

\begin{algorithm}[t]
\caption{Adversarial VAE-Based Defense}
\label{alg:adv_vae_denoising}
\footnotesize
\KwIn{Clean RSSI inputs $\{\mathbf{x}_{u,S}\}_{u=1}^N$,\\ 
 Simplified DNN function $f(\cdot;\boldsymbol{\theta})$,\\
Perturbation bound $\epsilon$,\\
Randomly initialized encoder $\mathscr{E}_{\boldsymbol{\Phi}}$, decoder $\mathscr{D}_{\boldsymbol{\Theta}}$.}
\KwOut{Denoised inputs $\hat{\mathbf{x}}_{\text{denoised}}$, predicted beams $\hat{y}$}

\BlankLine
\textbf{Phase 1: Adversarial Data Generation} \;
\For{$u = 1$ to $N$}{
    $\mathbf{g}_u \gets \nabla_{\mathbf{x}_{u,S}} \mathcal{L}\!\left(f(\mathbf{x}_{u,S}; \boldsymbol{\theta})\right)$ \;
    $\boldsymbol{\delta}_{u,S} \gets \epsilon \cdot \frac{\mathbf{g}_u}{\|\mathbf{g}_u\|_2}$ \;
    $\mathbf{x}_{u,S}^{\mathrm{adv}} \gets \mathbf{x}_{u,S} + \boldsymbol{\delta}_{u,S}$ \;
}

\BlankLine
\textbf{Phase 2: DVAE Training} \;
$\mathbf{X}_{\text{in}} \gets \text{concatenate}(\mathbf{x}_{u,S}, \mathbf{x}_{u,S}^{\mathrm{adv}})$, \\
$\mathbf{X}_{\text{target}} \gets \text{concatenate}(\mathbf{x}_{u,S}, \mathbf{x}_{u,S})$ \;

\ForEach{$\mathbf{x}_{u,S}, \mathbf{x'}_{u,S} \in \mathbf{X}_{\text{in}}, \mathbf{X}_{\text{target}}$}{
    $\hat{\mathbf{x}}_{u,S} \gets \mathscr{D}_{\boldsymbol{\Theta}}\!\left(\mathscr{E}_{\boldsymbol{\Phi}}(\mathbf{x}_{u,S})\right)$ \;
    $\mathcal{L}_{\text{DVAE}} \gets 
        \underbrace{\ell_r(\mathbf{x'}_{u,S}, \hat{\mathbf{x}}_{u,S})}_{\text{Reconstruction}} 
        + \underbrace{\ell_{\text{KL}}\!\big(\mathcal{N}(\boldsymbol{\mu}_{z_u}, \boldsymbol{\sigma}_{z_u}^2)\,\|\,\mathcal{N}(\mathbf{0}, \mathbf{I})\big)}_{D_\text{KL divergence}}$ \;
    Update parameters $(\boldsymbol{\Phi}, \boldsymbol{\Theta})$ using $\nabla_{(\boldsymbol{\Phi}, \boldsymbol{\Theta})}\mathcal{L}_{\text{DVAE}}$ \;
}

\BlankLine
\textbf{Phase 3: Inference and Denoising} \;
Given test input $\mathbf{x}_{u,S}^{\text{test}}$, \\
$\hat{\mathbf{x}}_{u,S}^{\text{denoised}} \gets 
    \mathscr{D}_{\boldsymbol{\Theta}}\!\left(\mathscr{E}_{\boldsymbol{\Phi}}(\mathbf{x}_{u,S}^{\text{test}})\right)$ \;
$\hat{y} \gets \arg\max_j f'_j(\hat{\mathbf{x}}_{u,S}^{\text{denoised}}; \boldsymbol{\theta})$ \;

\end{algorithm}

\subsection{End-to-end Fine Tuning}\label{fine-tune-BAE}
{As illustrated in \Large \textcircled{\footnotesize{T7}} \normalsize in Fig. \ref{proposed_flow}}, once the DVAE model is trained, the reconstructed denoised RSSI representations can be used to retrain the downstream beam classifier, further enhancing robustness. To achieve this, we also consider fine-tuning the encoder $\mathscr{E}_{\boldsymbol{\Phi}}$, decoder $\mathscr{D}_{\boldsymbol{\Theta}}$, and classifier $f(\cdot;\boldsymbol{\theta})$  in an end-to-end manner to optimize the combined loss
\begin{equation}\label{eq:e2e_loss}
\mathcal{L}_{\text{End-to-End}} = \mathcal{L}_{\text{DVAE}} +  \,{\mathcal{L}_{\mathrm{CE}}},
\end{equation}
This joint optimization enables the simplified DNN classifier to adapt to denoised latent features, further improving accuracy over the original classifier, as detailed in Section \ref{performance_comp}.

\section{Performance Evaluation}\label{sec:simulation}
In this section, we provide details of the simulation setup, DL model architecture, and discussion of the results.
\subsection{Simulation Setup}

 We simulate mmWave BS-UE communication in an urban setting representative of downtown Boston, using the  Boston5G scenario from DeepMIMO~\cite{alkhateeb2019deepmimo}, The BS  is equipped with a 32-element ULA at 15\,m height facing the negative $y$-axis, and single-antenna UEs at 2\,m. The $200\,\mathrm{m} \times 230\,\mathrm{m}$ area is discretized with $0.37\,\mathrm{m}$ UE spacing. Channel vectors $\mathbf{h} \in \mathbb{C}^{32 \times 1}$ are normalized by their maximum absolute value to improve training stability. The BS performs analog beamforming with $M_\mathrm{w}= 32$ sensing beams using a $N_{\mathrm{BS}}$-DFT codebook, whereas, for the narrow beam codebook $\mathbf{W}$, we use an O-DFT  with oversampling  factor of 4 to get a total of 128 narrow beams. 
 In all experiments, 70\% of the data is allocated for training, 10\% for validation, and the remaining 20\% for testing. Other simulation parameters are summarized in Table \ref{tab:combined-hyperparams}, unless otherwise specified.   


\begin{table}[!t]
\centering
\caption{\textsc{Simulation parameters}} 
\label{tab:combined-hyperparams}
\begin{adjustbox}{width=0.53\columnwidth,center}
\footnotesize
\begin{tabular}{|l|l|} 
\hline
\multicolumn{2}{|c|}{\textbf{Channel Generation Parameters}} \\ \hline
Active BS                   & 1                                 \\ \hline
BS transmit power           & 30 dBm                            \\ \hline
Active users                & 1--622                            \\ \hline
Antenna configuration       & $(1,32,1)$ (ULA)                  \\ \hline
Carrier frequency           & 28 GHz                            \\ \hline
System bandwidth            & 500 MHz                           \\ \hline
Antenna spacing             & $0.5\lambda$                      \\ \hline
$D_{\text{train}}$ size     & 57,144                             \\ \hline
\end{tabular}
\end{adjustbox}
\vspace{-5pt}
\end{table}

\begin{table}[!t]
\centering
\caption{\textsc{ Model architectures of simplified DNN at the BS and surrogate CNN for black box Attack}}
\label{tab:dnn_cnn_arch}
\begin{adjustbox}{width=1\columnwidth,center}
\begin{tabular}{|c|c|c|c|c|}
\hline
\multirow{2}{*}{\textbf{Layer}} & \multicolumn{2}{c|}{\textbf{Simplified DNN}} & \multicolumn{2}{c|}{\textbf{Surrogate CNN for {Black-box Attack }}} \\ \cline{2-5}
 & \textbf{Units} & \textbf{Activation} & \textbf{No. of Filters / Layer Type} & \textbf{Activation} \\ \hline
Input & $\widetilde{M}_{\mathrm{w}}$ & - & $\widetilde{M}_{\mathrm{w}}$ & - \\ \hline
Layer 1 & 128 & ReLU & 128 / (Conv1D, $k=3$, pad=1) & ReLU \\ \hline
Layer 2 & 64 & ReLU & 64 / (Conv1D, $k=3$, pad=1) & ReLU \\ \hline
Layer 3 & 32 & ReLU & 32 / (Conv1D, $k=3$, pad=1) & ReLU \\ \hline
Global Pooling & - & - & AdaptiveAvgPool1D(1) & - \\ \hline
Flatten & - & - & Flatten & - \\ \hline
Output & $Y = 128$ & Linear & FC(128) / Linear & Linear \\ \hline
\end{tabular}
\end{adjustbox}
\end{table}

\begin{table}[!t]
\vspace{-1pt}
\centering
\caption{\textsc{DVAE model architecture }}
\label{tab:vae_arch}
\begin{adjustbox}{width=0.53\columnwidth,center}
\footnotesize
\begin{tabular}{|c|c|c|}
\hline
Layer & Units & Activation Function \\ \hline
Input & $\widetilde{M}_{\mathrm{w}}$ & - \\ \hline
Encoder Layer 1 & 32 & ReLU \\ \hline
Encoder Layer 2 & 16 & ReLU \\ \hline
Latent Layer & 8 & - \\ \hline
Decoder Layer 1 & 16 & ReLU \\ \hline
Decoder Layer 2 & 32 & ReLU \\ \hline
Output & $\widetilde{M}_{\mathrm{w}}$ & Linear \\ \hline
\end{tabular}
\end{adjustbox}
\vspace{-10pt}
\end{table}


\subsection{Deep Learning Model Architectures} 

The simplified DNN at the BS is a fully connected network with three hidden layers, taking RSSI values over the selected $\widetilde{M}_{\mathrm{w}}$ beams as input, and its output layer consists of $Y=128$ neurons corresponding to the O-DFT indices. The adversary performs semi-white-box attacks using gradients of this DNN. For black-box attacks, a surrogate convolutional neural network (CNN) with three Conv1D layers is used to capture local correlations, also outputting a $128$-dimensional vector. Both the simplified DNN at the BS and the surrogate CNN at the adversary are trained for 100 epochs with batch size 256 using the selected RSSI inputs $\widetilde{M}_{\mathrm{w}}$ and cross-entropy loss. For denoising, a symmetric DVAE with a fully connected encoder, an 8-dimensional latent space, and a decoder is used; reconstruction is prioritized via $\alpha=0.9$ and $\beta=0.5$, and it is trained for 200 epochs with batch size 500. All models are optimized using Adam with a learning rate of $10^{-4}$. Architectural details are provided in Table~\ref{tab:dnn_cnn_arch} and Table~\ref{tab:vae_arch}.

\subsection{ Baselines for Performance Comparison}
For beam selection performance comparison, we consider the following benchmark methods:

\begin{enumerate}
\item \textbf{SVD perfect channel} based on the singular value
decomposition (SVD) of perfectly known channels under
$3$-bit quantized phase shifter. 


\item \textbf{Exhaustive search O-DFT} (with oversampling factor of $\times$4):   The BS exhaustively sweeps
the O-DFT codebook to scan $ N_{\mathrm{BS}} \times$4 narrow beams, and selects
the beam with the highest RSSI value for downlink transmission. 


\item \textbf{SHAP feature selection (FS)} uses SHAP method from XAI to rank RSSI contributions across candidate beams \cite{KHAN_TWC}. The BS then sweeps only the top-ranked wide beams, reducing overall beam sweeping.

{{\item \textbf{2-Tier Hierarchical Beam Search:} The BS employs a two-tier hierarchical codebook \cite{BA6} , where the tier-1 codebook comprises multiple wide beams, and the tier-2 codebook includes all the narrow beams from the adopted O-DFT codebook of size $Y$. Each narrow beam in the tier-2 codebook is  a child beam associated with a wide beam from the tier-1 codebook.
}

{\item \textbf{Fixed Beam Selection:} Based on the adaptation of the method proposed in \cite{jalali2024fast}, the BS  selects a fixed subset of RSSI values collected over a subset of narrow beams from the O-DFT codebook to train a CNN to associate the correct/best-oriented beam between the BS and UE. 
This learning-based approach uses the same number of inputs as the proposed approach, but does not account for the importance of input features (RSSIs) when selecting the input subset for model training.}}

\end{enumerate}

{To evaluate the effectiveness of the proposed DVAE-based defense methodology, we compare it with (i) defensive distillation \cite{AD1} (ii) No attack (iii) No defense (iii) DVAE (adversarial) and (iv) DVAE (fine-tuned) as detailed below:

\begin{itemize}

    \item \textbf{No attack} case serves as a baseline, representing inherent DNN misclassifications independent of perturbations.

    \item \textbf{No defense} case referring to the performance of the simplified DNN-based beam classifier without any defense in place to mitigate the attack.
    

    \item \textbf{Defensive distillation}, where a baseline model, implemented as the simplified DNN operating on the $\widetilde{M}_{\mathrm{w}}$ inputs, is first trained with a high softmax temperature $T$ to generate softened probabilities for the optimal beams. These soft targets train a distilled DNN of the same architecture on the original inputs using a combined loss of soft predictions and true labels. {This process smooths decision boundaries and enhances robustness against gradient-based adversarial attacks \cite{AD1}.} During inference, the distilled model runs at the BS with $T=1$.

    \item \textbf{Proposed DVAE (adversarial)}, where the DVAE processes the adversarial samples to reconstruct their clean counterparts. {In the simulation results, three DVAE operating conditions are evaluated: DVAE (clean inputs), where unperturbed samples are processed by the DVAE; DVAE (black box attack), where adversarial inputs are generated without access to the target DNN’s gradient information; and DVAE (white-box attack), where the adversary has full access to the target DNN’s gradients.}
    

    \item \textbf{Proposed DVAE (fine-tuned)}, where the  simplified DNN classifier is fine-tuned on the reconstructed inputs by training the entire pipeline in an end-to-end manner, optimizing the joint loss function in \eqref{eq:e2e_loss}.

\end{itemize}

}

\subsection{Performance Metrics}
The following metrics are used for performance evaluation:
\begin{itemize}

\item \textbf{Top-k Accuracy:}
It represents the probability that the optimal beam index lies within the top $k$ predicted beams.

\item {\textbf{Beam Sweep Time} $(T_\mathrm{sweep})$:}
This denotes the time required to scan a set of candidate beams. The total IA duration is 
$T_\mathrm{IA} = T_\mathrm{sweep} + T_\mathrm{predict}$, 
where $T_\mathrm{predict}$ is the beam prediction time determined by the BS processing capability, and $T_\mathrm{sweep} = \widetilde{M}_{\mathrm{w}} t_s$ depends on the number of scanned beams $\widetilde{M}_{\mathrm{w}}$ and the per-beam scanning time $t_s$ \cite{FastIA}. Since $T_\mathrm{sweep} \gg T_\mathrm{predict}$, the beam sweep time is used as the primary comparison metric.

\item \textbf{Effective Spectral Efficiency:}
This measures the achievable data rate considering the overhead of IA procedures. It is defined as
\begin{equation}
\mathrm{SE} = \frac{T_\mathrm{frame} - T_\mathrm{IA}}{T_\mathrm{frame}} 
\log_2\!\left(1 + \mathrm{SNR}\right),
\end{equation}
where $T_\mathrm{frame}$ is the coherence frame duration. {The SE here is computed using the SNR of the top-$k$ predicted narrow beams after the beam alignment phase}.


{\item \textbf{Perturbation-to-Signal Ratio (PSR):}
This is defined as the ratio of perturbation power to the total received signal power \cite{attack_beam,AA2}.  For performance analysis in terms of accuracy versus PSR, the perturbation bound $\epsilon$ is computed from the RSSI vector as $\epsilon = \sqrt{\mathrm{PSR}} \;\|\mathbf{x}_u\|_2$. This bounded perturbation ensures that the attack power remains within the signal power range. As PSR increases, the likelihood of detecting the attack also rises.


}

\end{itemize}

\subsection{Performance Evaluation} \label{performance_comp}

\subsubsection{Analysis of Causal Beam Discovery}
To reduce dataset complexity while keeping only the most meaningful features, i.e, the causally relevant RSSI inputs, we utilize the open-source python implementation of DirectLiNGAM \cite{directlingam}. We validate the relevance of the causally selected features by comparing them against the SHAP-based feature selection approach, where a pre-trained DNN, trained on the full dataset $\mathcal{D}$ is used to rank the input features. {To extract the causal inputs, i.e., the number of wide beams to be swept, a fixed feature selection approach is adopted, referred to as Causal FS. Initial experiments show that the DNN classifier accuracy saturates as the number of input beams $\widetilde{M}_{\mathrm{w}}$ used for RSSI measurements increases. Specifically, the validation accuracy stabilizes at $\widetilde{M}_{\mathrm{w}} = 13$, and no further performance gains are observed with additional inputs. Therefore, $\widetilde{M}_{\mathrm{w}} = 13$ is selected for all simulations to provide a fair trade-off between accuracy and beam sweeping latency.}

Fig.~\ref{fig:DAG} illustrates the DAG constructed using the DirectLiNGAM algorithm for causal beam selection.  
Each node represents either an  input feature or the target beam index, and the directed edges denote  direct or indirect causal relevance determined using the direct parents and total connectivity criterion.  For \( \widetilde{M}_{\mathrm{w}}= 13,  \) the causal ordering based on Algorithm \ref{alg:CFS} results in the selection of features 
$x_{17}, x_{18}, x_{16}, x_{19}, x_{20}, x_{14}, x_{15}, x_{13}, x_{21}, x_{22}, x_{24}, x_{23}, \text{and } x_{10}$. 
These nodes (shown in blue) correspond to either the direct parents of the target in the causal structure, 
or the features contributing indirectly through total connectivity. 

\begin{figure*}[t]
    \centering
    \begin{minipage}[b]{0.32\textwidth}
        \centering
        \includegraphics[width=\linewidth]{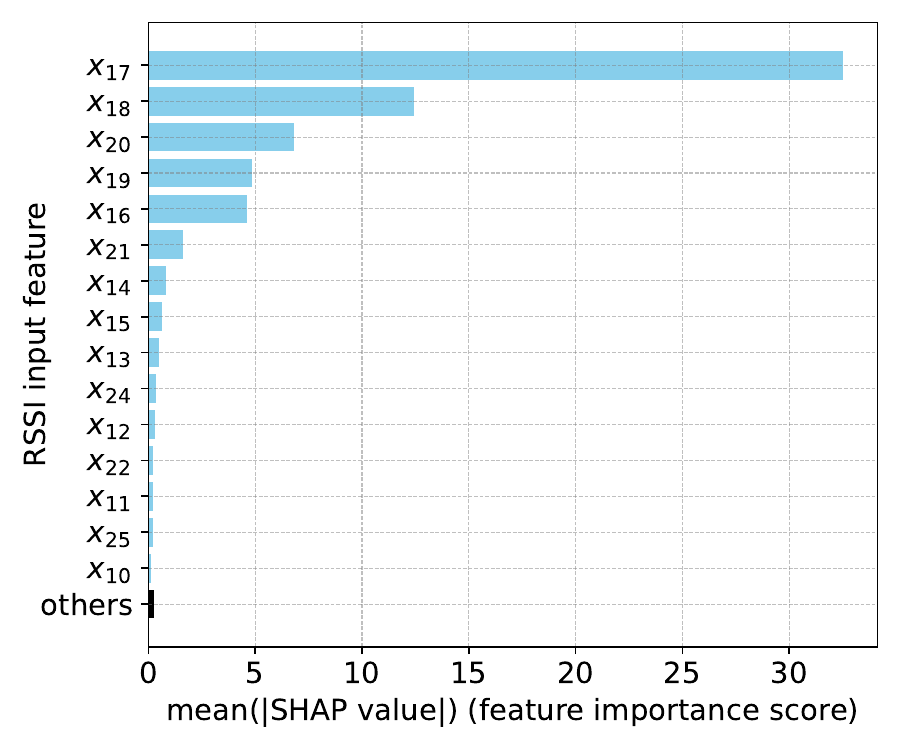}
        \caption{Feature importance rankings for the beam classifier using SHAP method.}
         \setlength{\abovecaptionskip}{-5pt}
        \label{fig:SHAP}
    \end{minipage}
    \hfill
    \begin{minipage}[b]{0.32\textwidth}
        \centering
        \includegraphics[width=\linewidth]{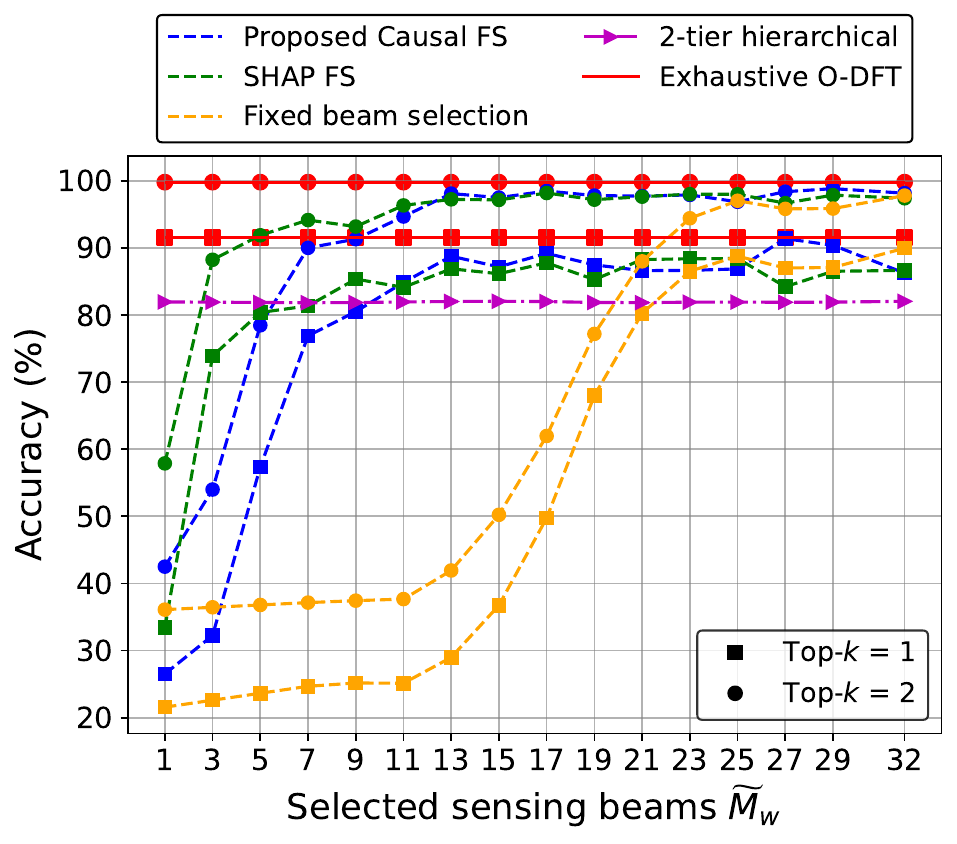}
        \caption{Top-k accuracy performance of different schemes vs. number of selected sensing beams/features.}
         \setlength{\abovecaptionskip}{-5pt}
        \label{fig:accuracy_vs_beams}
    \end{minipage}
    \hfill
    \begin{minipage}[b]{0.32\textwidth}
        \centering
        \includegraphics[width=\linewidth]{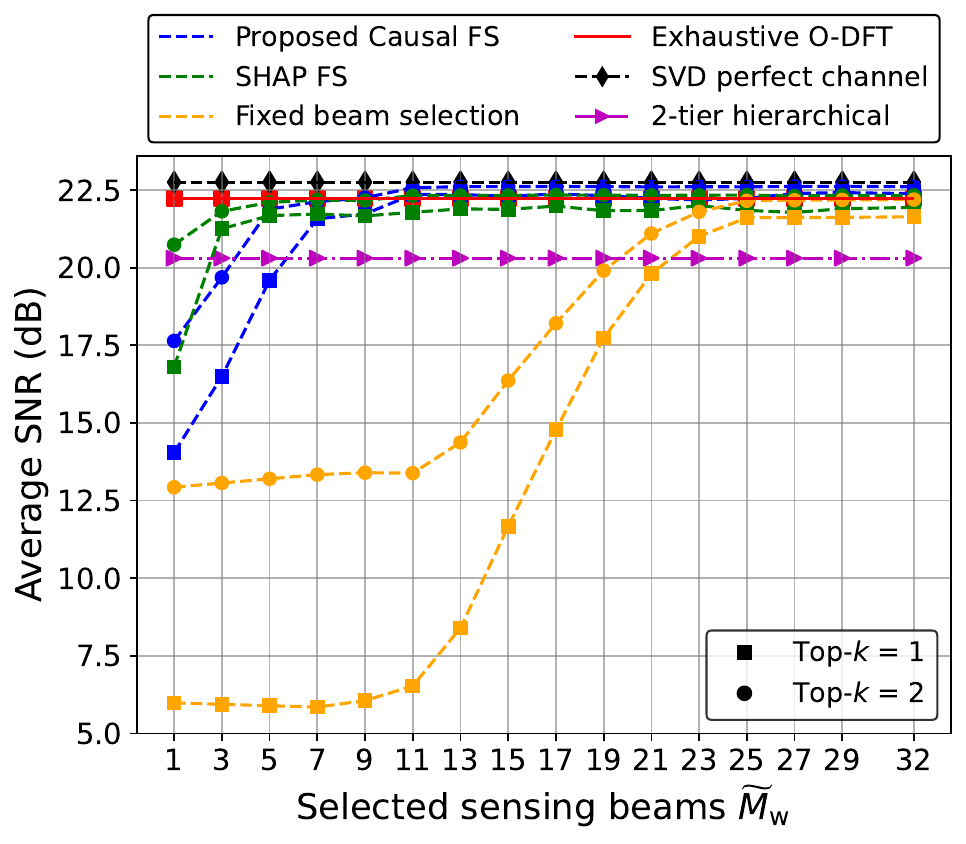}
        \caption{Average SNR performance of different schemes vs. number of selected sensing beams/features.}
         \setlength{\abovecaptionskip}{-5pt}
        \label{fig:SNR_vs_beams}
    \end{minipage}
     \setlength{\abovecaptionskip}{-5pt}
\end{figure*}

For comparison, Fig.~\ref{fig:SHAP} shows the feature importance rankings obtained using the SHAP method. The horizontal axis represents the average impact of each input on the classifier output, while the vertical axis orders the features in decreasing importance. {Notably, the causal features identified by DirectLiNGAM closely match the SHAP rankings, with 12 of the top 13 SHAP-ranked inputs coinciding with the proposed causal feature selection, validating the selection of these RSSI inputs for accurate and fast beam prediction.}  Note that the proposed causal selection approach only requires the observational dataset as input, effectively captures key predictive variables without the need for repeated model retraining, as required by SHAP based feature importance ranking.


\subsubsection{Analysis of Beam Alignment Accuracy } 
Fig.~\ref{fig:accuracy_vs_beams} illustrates the \textit{top-$k = \{1, 2\}$} accuracy as a function of the number of causality-based selected sensing input beam $\widetilde{M}_{\mathrm{w}}$. The exhaustive search approach yields the highest accuracy, primarily due to its reduced vulnerability to noise in beam measurements. Nonetheless, this gain comes at the cost of substantial training overhead and occasional alignment errors arising from noise-induced discrepancies.{ The hierarchical binary search performs the worst among the traditional beam sweeping-based baselines since the two-tier  search  introduces more layers and is thus more prone to error propagation; a suboptimal decision in any tier-1 may lead to an incorrect final beam selection due to noise or imperfect beam patterns. The importance of feature selection becomes evident from the performance of the fixed beam search approach, which treats all features/beams as equally important. It requires sweeping at least 23 sensing beams to achieve an accuracy of  86\%.} In contrast, the proposed Causal FS approach achieves 86\% beam alignment accuracy using only $\widetilde{M}_{\mathrm{w}} = 13$ sensing beams, and further improves to 97\% accuracy by exploring an additional $k = 2$ narrow beams. {This performance matches that obtained with SHAP FS approach, while avoiding model retraining for feature importance evaluation, despite negligible performance variation due to the input noise.}

\subsubsection{Analysis of Average SNR}


Fig.~\ref{fig:SNR_vs_beams} shows the average SNR of the proposed and benchmark schemes versus the number of selected features, $\widetilde{M}_{\mathrm{w}}$.  The exhaustive search attains near-optimal SNR comparable to the SVD-based solution by evaluating all beams in the O-DFT codebook. On the otherhand, the proposed {Causal FS} approach achieves similar SNR performance using only 13 of 32 causality-based selected sensing beams, demonstrating its ability to identify features most relevant to beam index selection. Unlike the SHAP-based approach, {Causal FS} attains comparable performance to the exhaustive search without requiring DNN pretraining, achieving $\approx 89\%$ reduction in beam sweeping overhead compared to exhaustive search while maintaining near-optimal alignment accuracy. {The degraded performance of fixed beam
search approach may be attributed to the fact that  randomly selecting a subset of  narrow beams for sweeping may overlook the optimal direction since it might not be included in
the measured set during training.}

\subsubsection{Analysis of Effective Spectral Efficiency} Fig.~\ref{fig:SE_vs_features} shows the effective spectral efficiency achieved by the proposed and benchmark schemes across varying numbers of selected features $\widetilde{M}_{\mathrm{w}}$. The exhaustive search method requires scanning $128$ narrow beams and has the worst effective spectral efficiency despite having superior average SNR performance. { The hierarchical search  require  $M_\mathrm{w} +\lceil Y / M_\mathrm{w} \rceil$  beam searches for $M_\mathrm{w}$ wide beams and $Y$ narrow beams, respectively, and therefore exhibit degraded effective spectral efficiency. } The proposed \textit{Causal FS} approach adaptively selects $\widetilde{M}_{\mathrm{w}}$ based on causal relevance, and  with just $\widetilde{M}_{\mathrm{w}} = 8$ beams, the proposed method outperforms the SVD-based solution and matches the performance of the \textit{SHAP FS} approach.

\subsubsection{Performance under semi-white box and black box attacks}

We begin by evaluating the learning efficacy of our proposed VAE in terms of reconstructing the original RSSI samples from their perturbed (adversarial) counterparts.

\begin{figure*}[t]
    \centering
    \begin{minipage}[b]{0.31\linewidth}
        \centering
        
        \includegraphics[width=\linewidth]{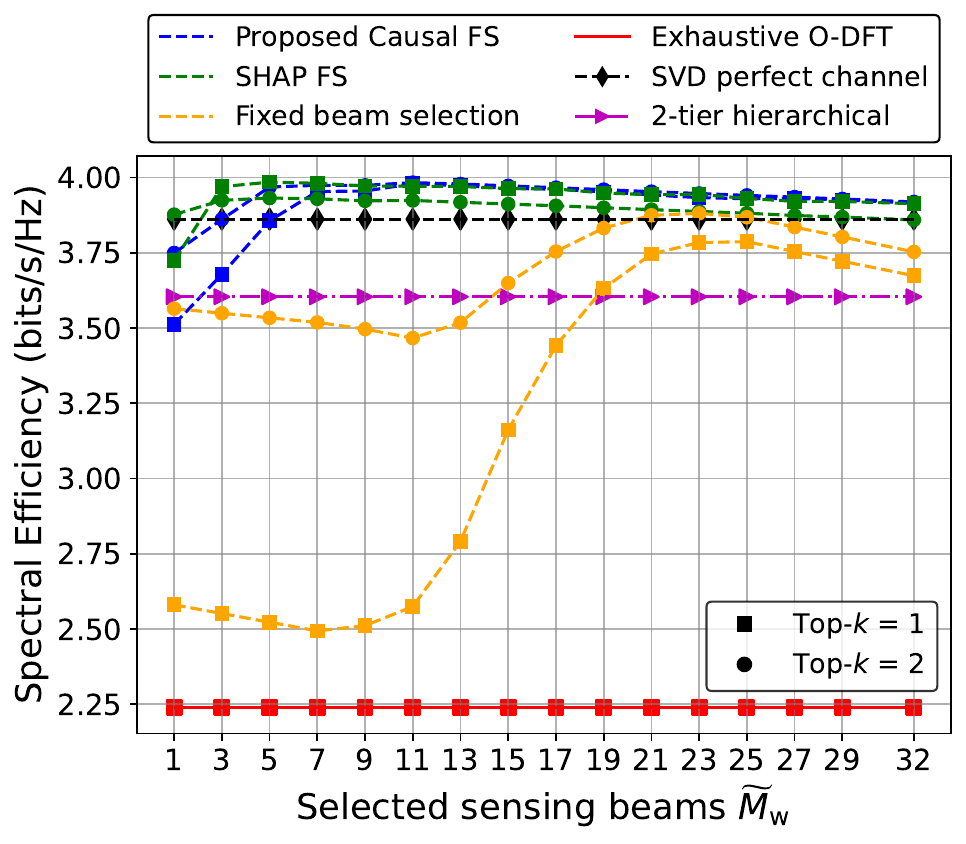} 
        \caption{Effective spectral efficiency vs. number of selected sensing beams/features for different approaches.}
         \setlength{\abovecaptionskip}{-10pt}
        \label{fig:SE_vs_features}
    \end{minipage}
    \hfill
    \begin{minipage}[b]{0.31\linewidth}
        \centering
        
       \includegraphics[width=\linewidth]{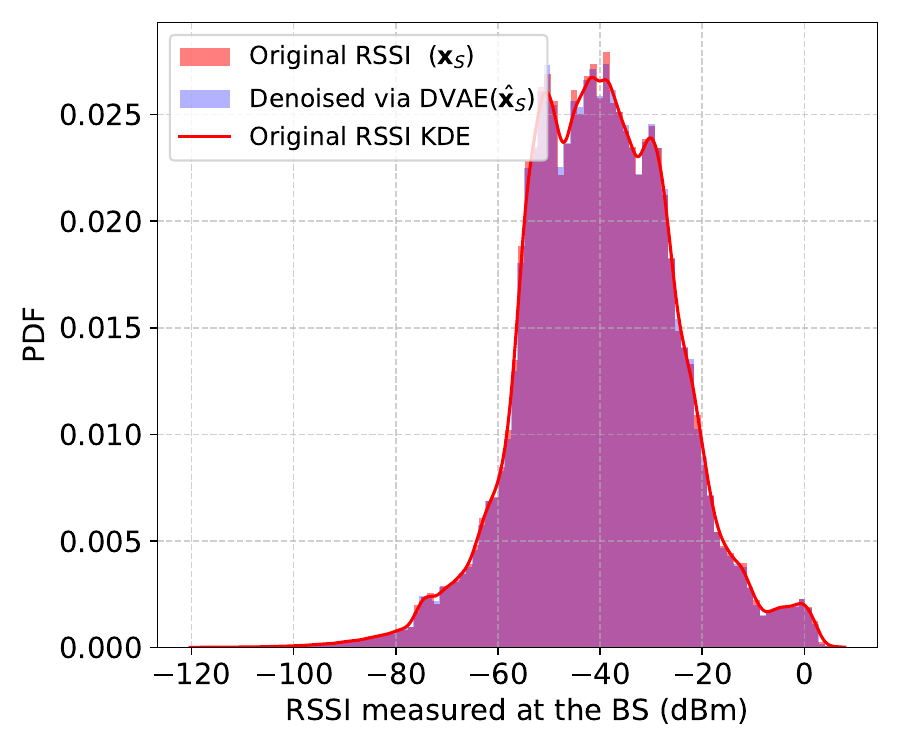}
        \caption{Denoising performance of the proposed DVAE model; the red curve shows the {kernel density estimation (KDE)} of the original RSSI values.}
         \setlength{\abovecaptionskip}{-10pt}
        \label{fig:denoising}
    \end{minipage}
    \hfill
    \begin{minipage}[b]{0.31\linewidth}
        \centering
        \includegraphics[width=\linewidth]{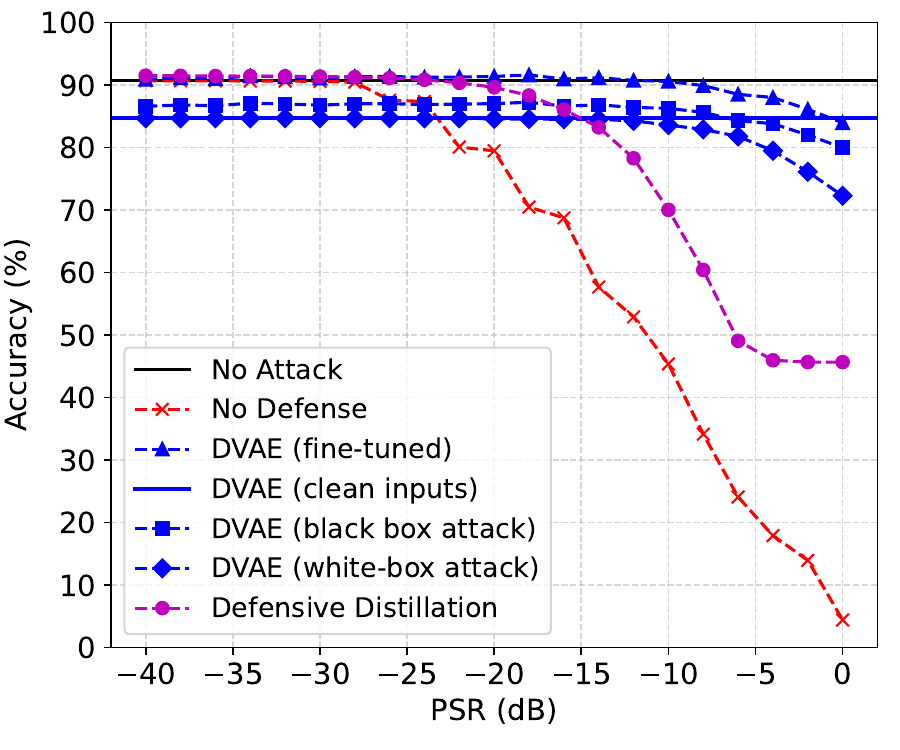}
        \caption{Top-1 beam prediction accuracy under white-box and black-box attacks with and without the proposed DVAE-based defense.}
        \setlength{\belowcaptionskip}{-10pt}
        \label{fig:vae-blackbox}
    \end{minipage}
\end{figure*}

Fig.~\ref{fig:denoising} illustrates the reconstruction performance of the proposed DVAE-based defense mechanism. 
Adversarial samples are generated under a white-box threat model with a perturbation-to-signal ratio (PSR) of $-10$~dB. 
The probability density function (PDF) depicts the distribution of the original unperturbed RSSI samples from the test dataset alongside their  reconstructed (denoised) outputs obtained after propagating the adversarial inputs through the trained DVAE model.  As shown, the DVAE effectively restores the RSSI values, achieving a mean squared reconstruction error below $0.1\%$.

Fig.~\ref{fig:vae-blackbox} illustrates the beam classifier’s Top-1  accuracy  under both  semi-white box and black-box adversarial attacks, with and without the proposed DVAE-based defense.  The semi-white box attacks significantly degrade beam selection accuracy, while black-box attacks are relatively less potent due to the mismatch between the surrogate and BS-side gradient directions. Defensive distillation remains vulnerable under strong perturbations, consistent with prior findings~\cite{AD1}. In contrast, the proposed DVAE-based defense effectively mitigates both attack types, maintaining $\approx85\%$ accuracy even at PSR~$=-5$~dB, and achieving performance close to the baseline model with no attack as well as retaining the  baseline
performance on clean (unperturbed) inputs intact. Furthermore, fine-tuning the simplified classifier at the BS in an end-to-end with denoised (DVAE) inputs further improves beam alignment accuracy, yielding up to a $5$--$8\%$ gain in accuracy across considered PSR levels.

Table~\ref{tab:defense_comparison_topk} summarizes the Top-k/1,2,3 beam classification accuracy under both white-box and black-box adversarial attacks across different defense schemes. {The absence of any defense mechanism results in a drastic degradation of Top-1 accuracy, dropping by more than 40\% and 80\%  for PSR of 10 dB and 0 dB, respectively,  compared to the no-attack baseline, highlighting the model’s high vulnerability to adversarial perturbations. Incorporating the proposed DVAE-based denoising substantially mitigates this effect, recovering up to 80--85\% Top-1 accuracy under strong attack scenarios i.e, PSR$\geq -20$ dB.} Further improvement is achieved when fine-tuning the beam classifier jointly with the denoised representations in an end-to-end manner, yielding accuracies within 2--3\% of the clean baseline and surpassing defensive distillation by over 15-20\% on average. The considered black-box attack is less potent than the semi-white box attack, it still reduces the DNN's performance and is the most likely attack an adversary will induce, given that adversaries typically have limited knowledge of the BS signal-processing modules. Further, the proposed  DVAE is trained purely on adversarial samples generated via the semi-white box threat model but provides more robustness against black-box adversarial attacks that the DVAE is not trained to defend.

\subsubsection{Computation and Time Complexity}

The online computational complexity of the proposed DVAE-assisted beam alignment framework is dominated by forward propagation through the DVAE and simplified DNN at the BS. For a fully connected layer with $U_\mathrm{in}$ inputs and $U_\mathrm{out}$ outputs, the per-sample cost is $\mathcal{O}(U_\mathrm{in} \cdot U_\mathrm{out})$. The simplified DNN  in Table~\ref{tab:dnn_cnn_arch}, with $L$ layers and maximum size $H_\mathrm{max}=128$, scales as $\mathcal{O}(L \cdot H_\mathrm{max}^2)$, while the DVAE in Table~\ref{tab:vae_arch}, with $S$ layers and maximum hidden layer size $H_\mathrm{DVAE}$, adds an additional complexity of $\mathcal{O}(S \cdot H_\mathrm{DVAE}^2)$. Thus, the total per-input complexity is $\mathcal{O}(S \cdot H_\mathrm{DVAE}^2 + L \cdot H_\mathrm{max}^2)$, which is efficient since $H_\mathrm{DVAE} \ll H_\mathrm{max}$. For instance in the proposed implementation, $H_\mathrm{DVAE} =32$, and $ H_\mathrm{max}=128$, respectively. Moreover, the DVAE training cost is independent of the simplified DNN model at the BS, enabling deployment with high-parameter DNNs for adversarial mitigation without significant offline overhead.

Lastly, we discuss the time complexity in terms of the beam sweeping time and the input feature selection time. The computational complexity of the proposed method is dominated by beam sweeping and feedback, both of which scale linearly with the number of swept beams. All $N_\mathrm{U}$ UEs simultaneously measure and report RSSIs over $\widetilde{M}_{\mathrm{w}}$ sensing beams during BS sweeping. If the BS additionally sweeps top-$k$ predicted beams per UE, the sweeping complexity becomes $\widetilde{M}_{\mathrm{w}} + N_\mathrm{U} \times k \cdot \mathbbm{1}_{\{k > 1\}}$, and feedback complexity is $N_\mathrm{U} \times \widetilde{M}_{\mathrm{w}} + N_\mathrm{U} \cdot \mathbbm{1}_{\{k > 1\}}$. For instance, 5G NR specifies $5\mathrm{ms}$ to sweep $64$ sensing beams, allowing $0.078 \, \mathrm{ms}$ per beam \cite{FastIA}. Compared to the exhaustive search requiring $\approx 10 \mathrm{ms}$ to sweep $128$ narrow beams, the proposed method completes beam sweeping in $\approx 1.01 \mathrm{ms}$ using only $13$ sensing beams, achieving near-exhaustive accuracy while reducing the sweeping time by $\approx 90\%$. Regarding the online prediction time, forward propagation through the simplified DNN requires $2.83 \times 10^{-3}$~s per input, while DVAE preprocessing adds an additional $0.65 \times 10^{-3}$~s, resulting in a total inference time of $3.48 \times 10^{-3}$~s per input, which remains practical for real-time deployment under the proposed defense framework. 

\begin{table*}[t]
\centering
\caption{\textsc{Top-$k$ ($k$=1,2,3) beam classification accuracy (\%) under white-box and black-box adversarial attacks for different defense schemes.}}
\renewcommand{\arraystretch}{1.1}
\setlength{\tabcolsep}{2pt}
\begin{tabular}{l|c|ccc|ccc|ccc|ccc|ccc}
\hline
\multirow{2}{*}{\textbf{Attack Type}} & 
\multirow{2}{*}{\textbf{Attack Strength (PSR)}} &
\multicolumn{3}{c|}{\textbf{No Attack}} &
\multicolumn{3}{c|}{\textbf{No Defense}} &
\multicolumn{3}{c|}{\textbf{DVAE (adversarial)}} &
\multicolumn{3}{c|}{\textbf{DVAE (fine-tuned)}} &
\multicolumn{3}{c}{\textbf{Defensive Distillation}} \\ \cline{3-17}
& & Top-1 & Top-2 & Top-3 & Top-1 & Top-2 & Top-3 & Top-1 & Top-2 & Top-3 & Top-1 & Top-2 & Top-3 & Top-1 & Top-2 & Top-3 \\ \hline

\textbf{White-box} & -20 dB & 90.3 & 96.1 & 98.2 & 79.5 & 81.8 & 85.1 & 85.2 & 90.7 & 95.9 & \textbf{90.2} & \textbf{95.1} & \textbf{97.2} & 88.4 & 93.1 & 96.7 \\
 & -10 dB & 90.3 & 96.1 & 98.2 & 45.5 & 49.9 & 53.6 & 83.2 & 89.4 & 95.0 & \textbf{87.4} & \textbf{90.0} & \textbf{93.8} & 69.9 & 74.5 & 79.6 \\
 & 0 dB & 90.3 & 96.1 & 98.2 & 4.7 & 15.3 & 18.8 & 72.4 & 86.3 & 91.7 & \textbf{85.4} & \textbf{88.1} & \textbf{92.2} & 45.3 & 50.6 & 53.9 \\ \hline

\textbf{Black-box} & -20 dB & 90.3 & 96.1 & 98.2 & 81.8 & 84.3 & 87.4 & 87.3 & 93.1 & 97.2 & \textbf{91.5} & \textbf{94.9} & \textbf{98.1} & 90.3 & 94.8 & 97.9 \\
 & -10 dB & 90.3 & 96.1 & 98.2 & 47.8 & 52.5 & 56.1 & 87.5 & 91.3 & 97.1 & \textbf{89.8} & \textbf{92.6} & \textbf{95.9} & 72.6 & 77.3 & 82.2 \\
 & 0 dB & 90.3 & 96.1 & 98.2 & 7.2 & 17.8 & 21.5 & 87.3 & 91.9 & 95.2 & \textbf{87.9} & \textbf{90.5} & \textbf{94.3} & 47.8 & 52.9 & 56.2 \\ \hline

\textbf{Average} &  & 90.3 & 96.1 & 98.2 & 44.4 & 50.3 & 53.8 & 83.8 & 90.4 & 95.3 & \textbf{88.9} & \textbf{92.0} & \textbf{95.1} & 69.1 & 75.5 & 77.7 \\ \hline
\end{tabular}
\label{tab:defense_comparison_topk}
\end{table*}

 The time complexity comparison for input selection is provided in Table~\ref{tab:time-comparison}. Here, we consider ${M}_{\mathrm{w}}=32$, i.e the full  input feature set with different number of sample sizes. The proposed DirectLiNGAM-based causal feature selection provides a significant computational advantage over SHAP. For instance, with 4,000 samples, DirectLiNGAM requires less than 15 s, whereas SHAP exceeds 12,600 s. This is because SHAP relies on repeated model retraining, with complexity growing exponentially with the number of features and linearly with the dataset size. In contrast, DirectLiNGAM exploits causal relationships without retraining, making it  scalable and efficient for large datasets. Since feature selection and the subsequent simplified DNN and DVAE training/fine-tuning are performed offline, higher computational costs are acceptable for real-time deployment. Moreover, unlike adversarial training and defensive distillation, which require retraining and depend on the   DNN classifier architecture, the DVAE training cost is independent of the underlying network architecture.


\begin{table}[!t]
\vspace{-5pt}
\centering
\caption{\textsc{Time complexity comparison of feature selection methods.}} 
\label{tab:time-comparison}
\begin{adjustbox}{width=0.6\columnwidth,center}
\footnotesize
\begin{tabular}{|l|c|c|} 
\hline
\textbf{Test Set Size} & \textbf{SHAP (s)} & \textbf{DirectLiNGaM (s)} \\ \hline
1000                         & 3189.08           & 7.85                      \\ \hline
2000                         & 6325.27           & 11.25                     \\ \hline
3000                         & 9515.65           & 13.83                     \\ \hline
4000                         & 12611.96          & 14.98                     \\ \hline
\end{tabular}
\end{adjustbox}
\vspace{-15pt}
\end{table}


\section{Conclusion}\label{sec:conclusion}
In this paper,  we have presented a causal AI-based framework for efficient and robust beam alignment in mmWave MIMO systems. By integrating causal discovery into the beam alignment process, our two-stage causal beam selection algorithm identifies a minimal set of relevant inputs, achieving performance comparable to conventional methods while significantly reducing input selection time and beam sweeping overhead. To enhance robustness against adversarial attacks, we have introduced a DVAE that effectively maps perturbed inputs to their clean states, maintaining high accuracy both under attack and in normal operation. The proposed approach  unifies interpretability, efficiency, and security in beam alignment. In future work, we aim to extend this causal AI framework to dynamically adapt to time-varying channels and mobile environments, further improving its practical applicability in real-world networks. 

\bibliographystyle{IEEEtran}\bibliography{IEEE_abr,XAI_bibliography}

\end{document}